\documentclass[11pt]{article}
\usepackage[T1]{fontenc}
\usepackage[utf8]{inputenc}
\usepackage{lmodern}
\usepackage{microtype}
\usepackage[margin=1in]{geometry}
\usepackage{amsmath,amssymb,amsthm}
\usepackage{graphicx}
\usepackage{booktabs}
\usepackage{tabularx}
\usepackage{array}
\usepackage{float}
\usepackage[numbers,sort&compress]{natbib}
\usepackage[hidelinks]{hyperref}
\usepackage{url}

\newcolumntype{C}{>{\centering\arraybackslash}X}
\newcommand{\paperkeywords}[1]{\par\vspace{0.5em}\noindent\textbf{Keywords:} #1}
\newcommand{\paperfigure}[2][]{%
  \IfFileExists{#2}{%
    \includegraphics[#1]{#2}%
  }{%
    \fbox{\parbox[c][2in][c]{0.9\linewidth}{\centering
      Missing figure file:\\[0.5em]\texttt{\detokenize{#2}}}}%
  }%
}

\title{NV-Tesseract-AD: Curriculum-Enhanced Conditional Diffusion for Multivariate Time-Series Anomaly Detection}

\author{%
Aditi Gautam\thanks{ORCID: \href{https://orcid.org/0009-0007-4172-2842}{0009-0007-4172-2842}. Corresponding author: \href{mailto:adgautam@nvidia.com}{adgautam@nvidia.com}},
Manikandan Ravikiran\thanks{ORCID: \href{https://orcid.org/0000-0002-2640-8528}{0000-0002-2640-8528}}, Saira Qureshi, Sena Fatma Ekiz,\\
Charan Jagwani, Mahmoud Elhadidy, and Sanjeev Roka\\[0.5em]
\normalsize NVIDIA
}
\date{}

\newcommand{\paperabstract}{We propose \textbf{NV-Tesseract-AD}, a curriculum learning--enhanced conditional
diffusion framework for multivariate time-series anomaly detection. Our central
hypothesis is that reconstruction difficulty can be explicitly structured along
two complementary axes---masking severity and diffusion noise level---and that
training from easier to harder tasks along both axes simultaneously stabilizes
optimization and improves residual discriminability. We integrate a Transformer-based
diffusion backbone with a three-stage dual curriculum over masking ratios and
diffusion timesteps, retaining anomalous samples throughout training and relying
on structured corruption to drive self-supervised representation learning. We evaluate NV-Tesseract-AD under two regimes: in-distribution training on the
TSB-UAD-M corpus and strict zero-shot transfer to ten EasyTSAD benchmark families
without per-dataset tuning. The dual-axis curriculum produces masking-invariant
reconstruction, confirmed by a near-zero loss gap between complementary masking
strategies within the first training stage, and yields approximately $3.7\times$
improvement in F1 and $56\%$ reduction in MSE relative to a non-curriculum
baseline. Zero-shot performance reaches F1\,$=$\,0.984 on NAB and F1\,$=$\,0.948
on NormA; weaker results on heterogeneous families are traced to unsupervised
threshold calibration rather than reconstruction quality, identifying calibration
as the dominant remaining detection bottleneck. Accelerated inference via
DPM-Solver++ reduces sampling cost by $3.5$--$5.7\times$ at the dataset level but
couples to thresholding behavior, establishing that sampler and threshold must be
validated jointly. Together, these findings demonstrate that aligning diffusion
denoising difficulty with curriculum learning principles provides a principled and
practical approach for robust anomaly detection in complex multivariate time series. }

\newcommand{\keywords}{time-series anomaly detection; diffusion models; denoising diffusion probabilistic models; curriculum learning; multivariate time series; missing data imputation; Transformer; zero-shot evaluation; thresholding}

\begin{document}
\maketitle

\begin{abstract}
\paperabstract
\end{abstract}
\paperkeywords{\keywords}
\vspace{1em}

\section{Introduction}
\label{sec:introduction}

Anomaly detection in multivariate time series is a foundational problem across
manufacturing, energy, and finance, where deviations from normal system behavior
can incur significant economic and safety costs. The objective is to identify
timestamps or windows whose observed dynamics deviate from learned normal
patterns--deviations that may indicate faults, attacks, or system degradations.
In practice, this task is inherently challenging: real-world streams are often
high-dimensional, non-stationary, and affected by structural missingness due to
sensor dropouts or sub-sampling, while anomaly labels are sparse or entirely
unavailable, precluding purely supervised approaches.

Deep learning has substantially advanced time-series anomaly detection by enabling
representation learning directly from raw signals. Recurrent networks,
convolutional architectures, and Transformers have demonstrated strong performance
across diverse benchmarks. Transformers, in particular, excel at modeling
long-range temporal structure and inter-variable dependencies through self-attention,
making them well-suited to the multivariate setting. Alongside discriminative
models, generative approaches have emerged as a compelling paradigm for anomaly
detection: by learning the distribution of normal behavior, generative models
identify anomalies as patterns that are poorly reconstructed or assigned low
likelihood under the learned model.

Denoising diffusion probabilistic models (DDPMs)~\cite{ho2020ddpm} have
recently attracted attention in time-series modeling. Diffusion models learn data
distributions through a forward process that progressively corrupts inputs with
noise and a learned reverse process that iteratively reconstructs clean signals.
When adapted to time series~\cite{pintilie2023,yang2023ddmt,xiao2023diffad,dou2025deanomaly},
diffusion models capture fine-grained temporal structure and multivariate
dependencies, naturally supporting anomaly detection via reconstruction residuals.
Their iterative refinement provides a more expressive generative mechanism than
single-pass autoencoders or variational approaches.

Despite these advantages, diffusion-based anomaly detection faces two structural
challenges. First, reconstruction difficulty is inherently heterogeneous. At low
noise levels, the model must perform fine-detail recovery of nearly-clean signals;
at high noise levels, it must reconstruct strongly corrupted inputs from sparse
contextual evidence. Uniform timestep sampling exposes the model to this full
difficulty spectrum from initialization, which can produce unstable gradient
dynamics and slow convergence before basic denoising competence is acquired.
Second, real-world multivariate streams introduce a masking dimension on top of
diffusion noise: when only a subset of observed channels serves as conditioning
context, the reconstruction task becomes harder as the masking fraction increases.
Standard diffusion training does not explicitly manage either source of difficulty.

Curriculum learning~\cite{bengio2009curriculum,soviany2022curriculum} provides a
principled framework for organizing training from easier to harder tasks, improving
convergence and generalization in a range of settings. In time-series modeling,
curricula have been applied to forecasting and anomaly detection by gradually
increasing prediction horizons or data volatility~\cite{koenecke2019curriculum,dube2024curriculum}.
In diffusion modeling for images, timestep curricula and adaptive noise schedules
have improved training stability~\cite{kim2024denoising}. However, the integration
of curriculum learning into diffusion-based anomaly detection for multivariate
time series--particularly the joint management of masking and noise difficulty
within a Transformer architecture--remains underexplored.

In this work, we introduce \textbf{NV-Tesseract-AD}, a conditional diffusion
framework for multivariate time-series anomaly detection trained under a
\emph{dual-axis curriculum}. We hypothesize that reconstruction difficulty can be
explicitly organized along two complementary axes: (i)~masking severity--the
fraction of observed values withheld as reconstruction targets--and (ii)~diffusion
noise level--the timestep, which controls how heavily the target entries are
corrupted. We design a staged easy-to-hard schedule that jointly increases both
masking ratio and expected timestep across three curriculum stages, progressively
expanding the reconstruction difficulty the model must handle. Unlike approaches
that filter anomalies from training data, we retain anomalous samples and rely on
structured corruption to guide representation learning under a self-supervised
conditional imputation objective.

Our experiments evaluate NV-Tesseract-AD under two complementary regimes:
in-distribution evaluation on the TSB-UAD-M corpus and strict zero-shot transfer
to the EasyTSAD benchmark suite across ten dataset families without per-dataset
tuning. Results show that the dual-axis curriculum produces masking-invariant
reconstruction--confirmed by a near-zero loss gap between complementary masking
strategies by the end of the first training stage--and yields ${\approx}3.7\times$
improvement in F1 and $56\%$ reduction in MSE relative to a non-curriculum
baseline. Zero-shot performance reaches F1 $= 0.984$ on NAB and F1 $= 0.948$ on
NormA, with weaker results on families exhibiting heterogeneous labeling
conventions, which our analysis traces to unsupervised threshold calibration
rather than reconstruction quality.

Our contributions are threefold:

\begin{itemize}
    \item We introduce NV-Tesseract-AD, a Transformer-based conditional diffusion
    architecture for multivariate time-series anomaly detection trained under
    dual curriculum schedules over masking severity and diffusion noise level.

    \item We formalize reconstruction difficulty along both axes via a joint
    difficulty measure and design a three-stage easy-to-hard training strategy
    that demonstrably stabilizes optimization and produces masking-invariant
    generalization.

    \item We empirically demonstrate improved reconstruction fidelity and
    consistent gains in anomaly detection performance across multiple benchmark
    families under a strict zero-shot protocol, and provide a systematic analysis
    of the threshold calibration bottleneck that limits deployed performance.
\end{itemize}

\section{Related Work}

\subsection{Time Series Anomaly Detection}

Time series anomaly detection (AD) has long been studied using statistical forecasting and residual thresholding approaches such as ARIMA and rule-based systems. While these approaches are often interpretable, they struggle to capture non-linear dependencies, multiscale temporal dynamics, and cross-sensor correlations in high-dimensional industrial streams, and they are brittle under non-stationarity and weak supervision. Deep learning approaches significantly advanced the field: recurrent architectures such as RNNs and LSTMs improved temporal modeling but suffered from vanishing gradients and limited scalability, while autoencoder-based methods (AE, VAE) learned representations of normality and detected anomalies via reconstruction errors.

Transformer-based models further reshaped AD by enabling long-range dependency modeling and scalable attention mechanisms. Architectures such as Anomaly Transformer, GTA, and RINAT introduced association discrepancy modeling, graph-augmented attention, and reversible normalization to mitigate distribution shifts. TranAD \cite{tranad2022} demonstrated strong performance using attention mechanisms to spot anomalies in multivariate time series. Despite strong results, these systems remain fundamentally discriminative: they typically learn decision boundaries or discrepancy scores directly and can be sensitive to training instability and train--test mismatch, particularly when anomalies are rare and labels are limited.

Recent work has increasingly turned to generative modeling, particularly diffusion models, to learn the distribution of normal behavior and quantify deviations probabilistically. Diffusion-based detectors treat anomalies as unlikely reconstructions under a learned stochastic generative process and have achieved state-of-the-art results on industrial benchmarks such as SWaT and NASA telemetry \cite{pintilie2023ddpm,diffad2023}. However, most existing systems optimize surrogate reconstruction likelihoods rather than detection-specific objectives and, importantly for this work, employ uniform training schedules that expose the model to a wide range of denoising difficulty throughout training without explicit structure.

\subsection{Generative Models for Anomaly Detection}

Generative approaches to time-series AD include GANs, VAEs, and their structured extensions. GAN-based methods generate synthetic normal patterns but are prone to instability and mode collapse, while hierarchical VAEs and factorized latent-variable models improve interpretability and scalability. A recurring theme across these generative families is that anomaly detection performance is closely tied to the quality and robustness of learned normal reconstructions, which becomes challenging in multivariate, non-stationary settings.

Diffusion models have recently emerged as a particularly powerful generative family for time series, supporting forecasting, imputation, change-point detection, and AD in a unified probabilistic framework. Early diffusion detectors learned multivariate normal distributions and scored anomalies via reconstruction likelihoods, outperforming AE baselines \cite{pintilie2023ddpm}. Subsequent advances incorporated conditional diffusion \cite{diffad2023}, dynamic decomposition \cite{d3r2023}, spatio-temporal graph encoders \cite{difftad2024}, and latent diffusion pipelines for industrial monitoring.

Notably, decomposition-based approaches such as D3R \cite{d3r2023} separated trend and residual components prior to diffusion-based reconstruction, improving robustness to concept drift, while graph-based diffusion models captured spatial correlations in traffic and sensor grids \cite{difftad2024}. TimeADDM \cite{timeaddm2024} introduced a score-based diffusion model that first encodes sequences with recurrent networks to capture global temporal context, then applies diffusion denoising on latent representations, computing multi-scale reconstruction errors as anomaly scores. AnoDDPM \cite{anoddpm2024} trains DDPMs on only healthy time-series and identifies anomalies via likelihood scoring, achieving high recall on aerospace sensor faults. Collectively, these systems highlight diffusion's ability to model complex temporal manifolds. Nevertheless, they typically inherit standard diffusion training practices (e.g., fixed corruption processes and uniform timestep sampling), leaving denoising difficulty largely unmanaged during optimization.

\subsection{Diffusion Models for Time Series}

Diffusion models were first introduced for vision but have rapidly expanded to time-series analysis, supporting probabilistic forecasting, imputation, anomaly detection, and industrial reliability modeling. Early work such as TimeGrad \cite{timegrad2021} introduced autoregressive diffusion for multivariate forecasting, while later models pursued non-autoregressive generation using Transformer backbones and sparse attention mechanisms. DiffInformer \cite{diffinformer2026} combines a Transformer (Informer with ProbSparse self-attention) and conditional diffusion to generate long-horizon forecasts in a single step, achieving improved accuracy on multi-step forecasting benchmarks while cutting computation versus standard Transformer models. Multi-resolution diffusion networks \cite{gao2023multiresolution} and non-stationary noise adaptation \cite{wang2024nsdiff} further improved long-horizon forecasting and calibration by learning coarse-to-fine time series structure and adapting noise perturbation over time to capture time-varying uncertainty.

For imputation, conditional diffusion models such as CSDI \cite{tashiro2021csdi} and state-space augmented diffusion systems \cite{sssd2022} demonstrated superior uncertainty-aware reconstruction of missing segments, including deployment in large-scale AIOps pipelines for cloud-failure detection \cite{yang2023aiops}. CSDI first demonstrated conditional score-based diffusion for imputation, treating known values as conditioning and learning to sample missing values via denoising steps, achieving high-quality imputation on clinical and environmental datasets. SSSD (Structured State Space Diffusion) integrates state-space models (S4) into the diffusion network for imputation and forecasting, exploiting long memory via state-space layers. MIDM \cite{midm2023} uses multivariate conditional diffusion to jointly impute missing values across correlated sensors, while PriSTI \cite{pristi2023} introduces diffusion for spatio-temporal imputation, capturing both temporal dynamics and spatial correlations in traffic data. In AD, diffusion has been combined with recurrent encoders, Transformers, decomposition modules, and graph neural networks to capture multi-scale structure and contextual dependencies.

Architecturally, diffusion models for time series increasingly replace convolutional denoisers with sequence-aware backbones, including Transformers, state-space layers, and cross-variable attention. TimeDiff \cite{timediff2023} uses a Transformer decoder conditioned on past sequence embeddings to produce multi-step forecasts via diffusion. Researchers have also proposed modified noise schedules and frequency-decomposition strategies to reflect non-stationarity, and diffusion-based Remaining Useful Life (RUL) prediction systems \cite{rul2024} for machinery monitoring that generate ensembles of degradation trajectories with calibrated uncertainty. Graph Diffusion Recurrent Networks combine GCN for inter-sensor relations with diffusion processes for temporal evolution \cite{graph_diffusion}. Despite these advances, most diffusion models for AD employ uniform sampling over diffusion steps and fixed corruption processes, leaving training difficulty unmanaged and creating an opening for structured optimization strategies that explicitly control denoising difficulty over the course of training.

\subsection{Curriculum Learning in Time Series}

Curriculum learning (CL) organizes training from easier to harder samples or tasks to improve stability and generalization. In time series, CL has been applied to forecasting via horizon curricula and scheduled sampling strategies that gradually expose models to longer prediction windows, yielding substantial error reductions and faster convergence. Teutsch and Mäder \cite{teutsch2022flipped} performed a comprehensive study of curriculum schedules for sequence-to-sequence LSTM forecasting, formalizing curricula along training horizon and iteration dimensions. Their "Flipped Classroom" training strategy, which combines increasing-horizon curriculum with probabilistic scheduled sampling, yielded up to 81\% lower NRMSE error compared to standard training, with models converging in fewer epochs and producing stabler long-range predictions. Spatio-temporal forecasting studies \cite{du2024sigspatial} similarly demonstrated that beginning with regular regimes and progressing to volatile patterns improves quantile accuracy, though overly tight curricula could lead to missed diversity.

For anomaly detection, curriculum strategies have been explored in industrial systems by ordering training windows based on reconstruction difficulty, window length, or data volatility. Dube \& Akar \cite{dube2024fruct} proposed a CL framework for industrial anomaly detection that orders training data from easy to hard via data subset and window length curricula. In their approach, an existing anomaly detector (MSCRED) is first trained on predictable segments with small windows and stable patterns, then incrementally on harder data. Experiments on the SWaT water treatment benchmark demonstrated that a three-stage data curriculum (33\%, 67\%, then full data in increasing difficulty) consistently improved anomaly detection F1-scores, outperforming no-curriculum training. The window-length curriculum (starting with short 10-step windows and gradually including longer 60-step windows) likewise boosted detection, as the model could first learn to reconstruct small fluctuations before tackling long-term dependencies. Interestingly, model-complexity curricula alone did not help unless combined with data curricula, suggesting that gradually increasing data complexity is more beneficial than simply starting with a weaker model.

\subsection{Curriculum Schedules in Diffusion Models}

Beyond dataset-level curricula, recent work has explored difficulty-aware training inside diffusion models themselves. Diffusion models traditionally sample timesteps uniformly, but recent analyses revealed that denoising difficulty varies substantially across noise levels. Kim et al. \cite{kim2025iclr} proposed timestep curricula that train models first on high-noise, coarse-structure reconstruction tasks before gradually incorporating low-noise, fine-detail steps, accelerating convergence and improving generation quality. They found that denoising at earlier timesteps (near the end of the diffusion process when input is only lightly noised) is actually more difficult than denoising heavily corrupted inputs, as the model must generate fine details rather than broad strokes. Their curriculum (high $t \rightarrow$ low $t$) led to better image generation performance and faster convergence ($\sim1.3\times$ faster) than uniform training by preventing the model from being overwhelmed early on by the hardest fine-detail denoising tasks.

Yi et al. \cite{yi2024cvpr} designed a diffusion time-step curriculum for one-image-to-3D generation (DTC123), using large diffusion timesteps (high noise) to learn coarse 3D geometry first, and only using small timesteps (low noise) later to refine high-frequency textures. Adaptive noise schedules further treat the corruption process as part of the curriculum. Hang \& Gu \cite{hang2024neurips} proposed ANT (Adaptive Noise schedule for Time series diffusion models), selecting a noise variance schedule based on dataset non-stationarity characteristics and reporting improved forecasting performance. Dataset-aware schedules and importance-weighted timestep sampling prioritized noise levels that provide more informative gradients \cite{hang2024importance}. Loss reweighting based on signal-to-noise ratios (Min-SNR strategy) \cite{laine2023minsnr} has similarly been shown to stabilize diffusion optimization by down-weighting loss at very noisy timesteps so the model focuses on moderate ones; related findings appear in \cite{choi2022,esser2024sd3}.

Curricula have also been applied along other axes, including progressive resolution, increasing mask ratios, and dual corruption schedules that anneal both noise magnitude and missing-value patterns. Progressive Growing of GANs \cite{karras2018progressive} introduced resolution curriculum for GAN training by gradually increasing output resolution. In masked autoencoders, progressive masking has been used as curriculum: Jar\v{c}a et al. \cite{jarca2024cbm} annealed mask ratios and biased masks to salient regions; Madan et al. \cite{madan2024clmae} introduced CL-MAE with a learnable masking module that transitions from cooperative to adversarial masks; and Lee et al. \cite{lee2022curriculum} applied curriculum masking in language modeling to improve rare concept learning. Related curriculum mechanisms have also been explored in RL settings \cite{tang2024neurips,eppe2018,lin2024prototype}.

\subsection{Gap and Motivation}
\label{subsec:gap}

Synthesizing the above, three specific gaps motivate this work.

\textbf{No unified framework integrates diffusion, Transformer architectures, and
curriculum learning for multivariate time-series anomaly detection.}
Prior work has studied each component in isolation: diffusion models for
time-series AD~\cite{pintilie2023ddpm,diffad2023,d3r2023,timeaddm2024},
Transformer-based temporal modeling~\cite{diffinformer2026,timediff2023,tranad2022},
and curriculum learning either for time-series
modeling~\cite{teutsch2022flipped,dube2024fruct} or for diffusion
training~\cite{kim2025iclr,hang2024neurips}. Existing diffusion--Transformer
hybrids largely focus on forecasting and omit curriculum mechanisms. Curriculum-enhanced
time-series models have been developed without generative diffusion objectives.
No system has unified all three within a single principled framework that is
evaluated under realistic non-stationarity and weak supervision.

\textbf{Existing curriculum approaches operate along a single difficulty axis,
and joint scheduling of masking and noise has not been systematically studied.}
Prior timestep curricula for diffusion~\cite{kim2025iclr,yi2024cvpr} vary only
noise level; prior data and window-length curricula for time-series
AD~\cite{dube2024fruct} vary only input complexity. Neither line of work considers
the interaction between masking severity and diffusion noise level as a
two-dimensional difficulty space, nor provides a theoretically grounded joint
difficulty measure that motivates the curriculum design. Yet both axes compound in
masked conditional diffusion: at high masking and high noise simultaneously,
the model must reconstruct strongly corrupted targets from sparse context, a
setting qualitatively harder than either axis alone.

\textbf{The contribution of threshold calibration to deployed performance gaps
is not systematically analyzed.}
Most existing diffusion-based AD systems optimize surrogate reconstruction
objectives and rely on post-hoc thresholding, without characterizing how much of
the gap between reconstruction quality and detection performance is attributable
to the thresholding mechanism versus the generative model
itself~\cite{pintilie2023ddpm,diffad2023}. This makes it difficult to identify
where future improvements will have the highest leverage. Motivated by these gaps, this paper proposes NV-Tesseract-AD: a conditional
diffusion framework with a Transformer backbone trained under a dual-axis
curriculum over masking severity and diffusion noise level, evaluated under a
strict zero-shot protocol, and accompanied by a systematic decomposition of the
reconstruction--detection performance gap.
\section{Methodology}
\label{sec:method}

\subsection{Method Overview}
\label{subsec:overview}

We propose \textbf{NV-Tesseract-AD}, a conditional diffusion framework for multivariate
time-series anomaly detection trained under a \emph{dual-axis curriculum}. The
central premise is that reconstruction difficulty in a masked diffusion setting is
not uniform across training iterations: it is determined jointly by (i) the
proportion of observed values withheld from the model (masking severity) and
(ii) the extent to which those withheld values are corrupted during the forward
diffusion pass (noise level). Conventional diffusion training exposes the model
to the full joint difficulty spectrum from initialization, which we argue creates
unstable optimization dynamics when learning from heterogeneous, non-stationary
industrial streams.

Instead, NV-Tesseract-AD structures training as an explicit easy-to-hard curriculum
over both axes simultaneously. Early in training, the model solves low-masking,
low-noise reconstruction tasks analogous to ``filling in a few missing pixels in a
slightly blurry photograph.'' As training progresses, the curriculum increases
both the fraction of withheld values and the noise severity, ultimately requiring
the model to reconstruct heavily masked, strongly corrupted inputs. We hypothesize 
and empirically confirm that this staged exposure (1) stabilizes gradient
dynamics in early training, (2) reduces residual variance under the learned
generative process, and (3) yields reconstruction residuals that are more
discriminative for downstream anomaly detection.
\begin{figure}[!htbp]
    \centering
    \paperfigure[
        width=0.99\linewidth,
        height=0.72\textheight,
        keepaspectratio
    ]{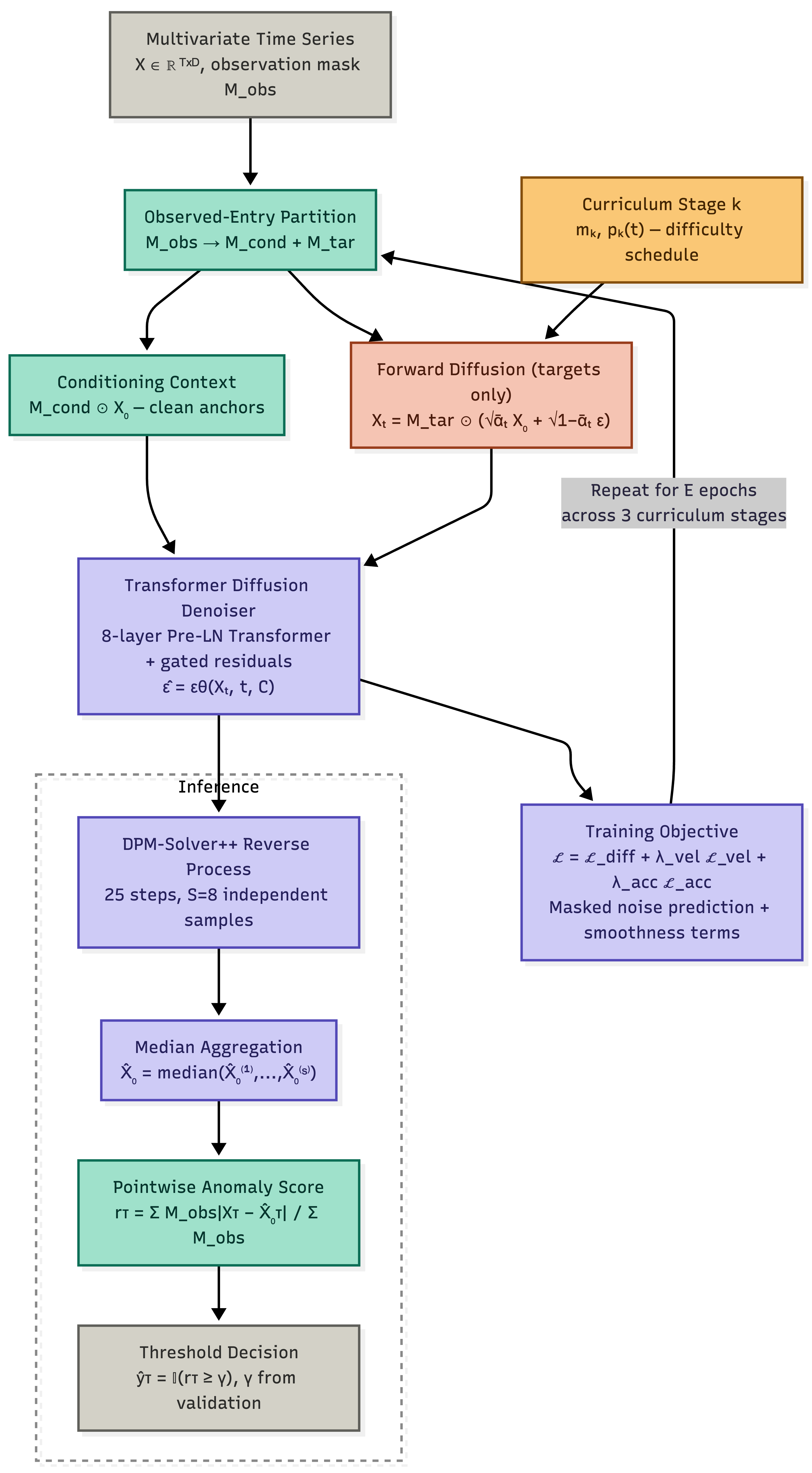}
    \caption{End-to-end pipeline of NV-Tesseract-AD.}
    \label{fig:overview}
\end{figure}

The method integrates five tightly coupled components, each described in
detail in subsequent subsections:

\begin{enumerate}
    \item \textbf{Conditional reconstruction formulation} (\S\ref{subsec:problem}).
    Observed entries in each training window are split into a conditioning set
    (revealed to the model as context) and a target set (withheld and used for
    supervision). Natural missingness is tracked separately and excluded from both
    conditioning and loss computation.

    \item \textbf{Masked conditional diffusion process} (\S\ref{subsec:diffusion}).
    A discrete-time DDPM forward process corrupts only the target entries, while
    conditioning entries remain intact as clean contextual anchors. The reverse
    process is parameterized by a learned denoising network.

    \item \textbf{Transformer-based diffusion denoiser} (\S\ref{subsec:transformer}).
    A residual Transformer with multi-head self-attention serves as the denoising
    backbone. It consumes corrupted target entries alongside clean conditioning
    context, mask indicators, diffusion-timestep embeddings, and curriculum-stage
    embeddings. Gated residual connections stabilize learning under high-noise
    inputs.

    \item \textbf{Dual-axis curriculum learning} (\S\ref{subsec:curriculum}).
    Training is divided into three stages of increasing difficulty. At each stage,
    both the masking ratio and the range of sampled diffusion timesteps expand,
    implementing a monotone schedule over a theoretically grounded joint difficulty
    measure.

    \item \textbf{Residual-based anomaly scoring} (\S\ref{subsec:inference}).
    At inference time, the trained denoiser reconstructs held-out entries through
    the reverse diffusion process. Pointwise anomaly scores are computed as
    normalized reconstruction residuals over observed variables, and binary
    decisions are obtained by thresholding on held-out validation data.
\end{enumerate}

\noindent Figure~\ref{fig:overview} illustrates the complete pipeline from a
training window through curriculum-controlled corruption and denoising to
anomaly scoring.

\subsection{Problem Setup}
\label{subsec:problem}

\paragraph{Time series and observation mask.}
Let $\mathbf{X} \in \mathbb{R}^{T \times D}$ denote a multivariate time series
window of length $T$ with $D$ sensor channels. In practice, industrial streams
frequently contain structural missingness due to sensor dropouts, communication
failures, or intentional sub-sampling. We capture this via an observation mask
$\mathbf{M}_{\mathrm{obs}} \in \{0,1\}^{T \times D}$, where
$\mathbf{M}_{\mathrm{obs},\tau,d}=1$ if entry $(\tau,d)$ is available and
$\mathbf{M}_{\mathrm{obs},\tau,d}=0$ if it is structurally missing. Crucially,
we never treat structural missingness as an anomaly indicator: absent entries are
excluded from both the diffusion loss and the downstream anomaly score. This
design choice prevents the model from conflating sensor dropout with fault
signatures, which is particularly important in heterogeneous industrial datasets
where missingness rates vary substantially across files.

\paragraph{Self-supervised training partition.}
For each training window, we further partition the set of observed entries
$\mathbf{M}_{\mathrm{obs}}$ into two disjoint complementary subsets:
\begin{equation}
\mathbf{M}_{\mathrm{obs}} = \mathbf{M}_{\mathrm{cond}} + \mathbf{M}_{\mathrm{tar}},
\qquad
\mathbf{M}_{\mathrm{cond}} \odot \mathbf{M}_{\mathrm{tar}} = \mathbf{0},
\label{eq:partition}
\end{equation}
where $\mathbf{M}_{\mathrm{cond}}$ is the \emph{conditioning mask} (entries
revealed to the model as context) and $\mathbf{M}_{\mathrm{tar}}$ is the
\emph{target mask} (entries withheld and used for supervision). The masking ratio
\begin{equation}
m = \frac{\sum_{\tau,d}\mathbf{M}_{\mathrm{tar},\tau,d}}
         {\sum_{\tau,d}\mathbf{M}_{\mathrm{obs},\tau,d}} \in [0,1]
\label{eq:maskratio}
\end{equation}
quantifies the fraction of observed entries withheld per window. This partition
creates a self-supervised conditional imputation objective: the model is trained
to infer withheld observations from partial context, without ever accessing
anomaly labels.

\paragraph{Learning objective (informal).}
Rather than parameterizing the conditional density $p_\theta(\mathbf{X}_0 \mid
\mathbf{M}_{\mathrm{cond}} \odot \mathbf{X}_0)$ in closed form which would
require intractable normalization over high-dimensional multivariate
distributions we adopt the denoising score-matching objective of DDPMs. Formally,
we learn a denoising network
\begin{equation}
\varepsilon_\theta(\mathbf{X}_t, t, \mathbf{C})
\label{eq:denoiser}
\end{equation}
that predicts the Gaussian noise added to the target entries at diffusion step
$t$, conditioned on contextual information $\mathbf{C}$ (defined in
\S\ref{subsec:transformer}). The reverse process defined by this denoiser
implicitly samples from the conditional distribution over clean target values.
Anomaly detection then follows by comparing observed values to their reconstructed
counterparts under this generative model.

\subsection{Masked Conditional Diffusion Reconstruction}
\label{subsec:diffusion}

\subsubsection{Forward Diffusion Process}
\label{subsubsec:forward}

We adopt a discrete-time DDPM \citep{ho2020ddpm} with $N=1000$ diffusion steps
and a cosine variance schedule $\{\beta_t\}_{t=1}^N$. Define the noise schedule
parameters:
\begin{equation}
\alpha_t = 1-\beta_t,
\qquad
\bar{\alpha}_t = \prod_{i=1}^{t}\alpha_i.
\label{eq:alphabar}
\end{equation}
The quantity $\bar{\alpha}_t$ controls the signal-to-noise ratio at step $t$:
as $t \to N$, $\bar{\alpha}_t \to 0$ and the data is dominated by Gaussian noise;
as $t \to 0$, $\bar{\alpha}_t \to 1$ and the data is nearly clean. A cosine
schedule \citep{nichol2021improved} maintains higher signal fidelity at low timesteps
compared to the linear schedule of the original DDPM, which we found particularly
beneficial for time-series data where fine temporal structure must be preserved
during easy curriculum stages.

In standard unconditional diffusion, the forward marginal at step $t$ is:
\begin{equation}
q(\mathbf{X}_t \mid \mathbf{X}_0)
= \mathcal{N}\!\left(
\sqrt{\bar{\alpha}_t}\,\mathbf{X}_0,\,
(1-\bar{\alpha}_t)\mathbf{I}
\right).
\label{eq:forward_marginal}
\end{equation}
Applying this directly to all entries would corrupt the conditioning context,
destroying the information that the denoiser relies on. We therefore apply
corruption \emph{selectively} to the target entries only.

\subsubsection{Masked Conditional Corruption}
\label{subsubsec:corruption}

Let $\boldsymbol{\varepsilon} \sim \mathcal{N}(\mathbf{0},\mathbf{I})$ be an
independent noise sample drawn per training iteration. We define the
\emph{masked corrupted input} at diffusion step $t$ as:
\begin{equation}
\mathbf{X}_t
= \underbrace{\mathbf{M}_{\mathrm{cond}} \odot \mathbf{X}_0}_{\text{clean context}}
+ \underbrace{\mathbf{M}_{\mathrm{tar}} \odot
  \!\left(\sqrt{\bar{\alpha}_t}\,\mathbf{X}_0
  + \sqrt{1-\bar{\alpha}_t}\,\boldsymbol{\varepsilon}\right)}_{\text{corrupted targets}}.
\label{eq:masked_corruption}
\end{equation}
This construction satisfies three design requirements:

\begin{enumerate}
    \item \textbf{Contextual integrity.} Entries in $\mathbf{M}_{\mathrm{cond}}$
    retain their clean observed values, providing the denoiser with an informative
    partial view of the time series at every noise level. This mimics the
    imputation setting at inference time, where some variables are fully observed
    and serve as anchors for reconstructing others.

    \item \textbf{Controlled target corruption.} Entries in $\mathbf{M}_{\mathrm{tar}}$
    are corrupted according to the standard DDPM forward marginal conditioned on
    $\bar{\alpha}_t$. The degree of corruption is monotonically controlled by
    the curriculum stage through the distribution $p_k(t)$ over timesteps
    (see \S\ref{subsubsec:schedule}), enabling explicit difficulty progression.

    \item \textbf{Structural missingness exclusion.} Entries where
    $\mathbf{M}_{\mathrm{obs},\tau,d}=0$ are excluded from $\mathbf{X}_t$, from
    the conditioning signal, and from the loss computation. The denoiser learns
    to operate robustly over incomplete observation patterns, which is essential
    for generalization to real-world multivariate streams.
\end{enumerate}

\subsubsection{Reverse Denoising and Clean Signal Recovery}
\label{subsubsec:reverse}

The denoising network $\varepsilon_\theta$ is trained to predict the Gaussian
noise $\boldsymbol{\varepsilon}$ that was added to the target entries:
\begin{equation}
\hat{\boldsymbol{\varepsilon}}
= \varepsilon_\theta(\mathbf{X}_t, t, \mathbf{C}),
\label{eq:denoise_pred}
\end{equation}
where the conditioning information $\mathbf{C}$ encompasses the full contextual
signal available at training time (detailed in \S\ref{subsec:transformer}).

Given the predicted noise, the clean estimate of all entries can be recovered
via the standard DDPM noise-prediction reparameterization:
\begin{equation}
\hat{\mathbf{X}}_0
= \frac{1}{\sqrt{\bar{\alpha}_t}}
  \left(\mathbf{X}_t - \sqrt{1-\bar{\alpha}_t}\;\hat{\boldsymbol{\varepsilon}}\right).
\label{eq:x0_estimate}
\end{equation}
In practice, only the target entries of $\hat{\mathbf{X}}_0$ are used for
supervision and anomaly scoring; conditioning entries are held fixed to their
observed values throughout the reverse process. At inference, the full reverse
Markov chain (or an accelerated solver; see \S\ref{subsec:inference}) is applied
iteratively from a sampled noise initial state $\mathbf{X}_N \sim \mathcal{N}(\mathbf{0},
\mathbf{I})$ at target positions, converging to a reconstruction $\hat{\mathbf{X}}_0$
that represents a sample from the model's learned conditional distribution over
clean target values.

\subsection{Transformer-Based Diffusion Denoiser}
\label{subsec:transformer}

\subsubsection{Input Representation and Token Construction}
\label{subsubsec:tokens}

The denoising network processes the multivariate time series as a sequence of
$T$ tokens, one per time step. Each token aggregates information from multiple
sources relevant to the denoising task. Concretely, the token at time step $\tau$
is constructed by channel-wise concatenation and subsequent linear projection of:

\begin{enumerate}
    \item \textbf{Corrupted series values.} The $D$-dimensional vector
    $\mathbf{X}_{t,\tau,\cdot}$, which equals the clean observation for
    conditioning entries and the noise-corrupted value for target entries.

    \item \textbf{Sinusoidal diffusion timestep embedding.} Following the
    approach of \citet{ho2020ddpm}, we encode the scalar diffusion step $t$
    as a $d_{\mathrm{emb}}$-dimensional sinusoidal embedding, then project it
    through two learned linear layers with GELU activations. The result is
    broadcast and added to every token, allowing the model to modulate its
    denoising behavior based on the current noise level.

    \item \textbf{Positional or temporal embedding.} A learned positional
    embedding for the time step $\tau$ within the window provides the model
    with temporal order information, which is critical for modeling sequential
    dependencies in time series.

    \item \textbf{Binary mask indicators.} Three binary scalar indicators per
    entry specify whether each variable at this time step belongs to
    $\mathbf{M}_{\mathrm{cond}}$, $\mathbf{M}_{\mathrm{tar}}$, or
    $\mathbf{M}_{\mathrm{obs}}^c$ (structurally missing). These signals prevent
    the denoiser from confusing observed contextual values with corrupted target
    values or structural absences.

    \item \textbf{Curriculum-stage embedding.} A learned embedding for the
    current curriculum stage $k \in \{1,2,3\}$, added to every token. This
    allows the model to develop stage-specific denoising strategies: at stage 1,
    the model can rely on abundant context and low noise; at stage 3, it must
    adopt a more global inference strategy under heavier corruption. Providing
    the stage index at test time (set to stage 3) ensures the model applies its
    most general denoising policy.
\end{enumerate}

\noindent The token embeddings are projected via a shared linear layer to a common
hidden dimension $H=256$, yielding the input sequence
$\mathbf{Z}^{(0)} \in \mathbb{R}^{T \times H}$.

\subsubsection{Residual Transformer Backbone}
\label{subsubsec:transformer_blocks}

The core of the denoiser is a stack of $L=8$ residual Transformer blocks with
Pre-Layer Normalization (Pre-LN) \cite{xiong2020layer}. The Pre-LN configuration
applies layer normalization \emph{before} the attention and feed-forward modules
rather than after their residual connections, which is known to produce more
stable gradient flow in deep architectures and was found empirically beneficial
during high-noise curriculum stages.

For block $\ell$, the update equations are:
\begin{align}
\mathbf{U}^{(\ell)}
&= \mathbf{Z}^{(\ell-1)}
   + \mathrm{MHA}\!\left(\mathrm{LN}\!\left(\mathbf{Z}^{(\ell-1)}\right)\right),
\label{eq:attn_block} \\
\mathbf{V}^{(\ell)}
&= \mathbf{U}^{(\ell)}
   + \mathrm{FFN}\!\left(\mathrm{LN}\!\left(\mathbf{U}^{(\ell)}\right)\right),
\label{eq:ffn_block}
\end{align}
where $\mathrm{MHA}(\cdot)$ is multi-head self-attention with 8 heads and
$\mathrm{FFN}(\cdot)$ is a position-wise feed-forward network with a $4\times$
hidden expansion ratio and GELU activation.

Self-attention is computed over the $T$ time-step tokens, allowing each position
to attend to every other position in the window. Formally, for a single attention
head with queries $\mathbf{Q}$, keys $\mathbf{K}$, and values $\mathbf{V}$ derived
from the normalized token sequence:
\begin{equation}
\mathrm{Attention}(\mathbf{Q},\mathbf{K},\mathbf{V})
= \mathrm{softmax}\!\left(\frac{\mathbf{Q}\mathbf{K}^\top}{\sqrt{d_k}}\right)\mathbf{V},
\label{eq:attention}
\end{equation}
where $d_k = H/n_{\mathrm{heads}}$ is the per-head key dimension. Full temporal
self-attention (without causal masking) allows the denoiser to leverage future
context when reconstructing each time step, which is appropriate for anomaly
detection where the full window is available at inference. This also enables the
model to capture long-range temporal dependencies that may be informative for
distinguishing anomalous patterns from smooth normal dynamics.

Rather than applying separate attention over the variable axis, NV-Tesseract-AD
uses a single per-time-step token that concatenates all $D$ channels. This design
allows cross-variable interactions to be modeled implicitly through the shared
hidden representations and the feed-forward layers, without introducing a quadratic
cost in $D$. This is particularly important given the variable dimensionality of
the training corpus, where files range from 2 to 64 channels after preprocessing.

\subsubsection{Gated Residual Refinement}
\label{subsubsec:gating}

After the standard Transformer update, we apply a gated residual refinement
step \citep{dauphin2017language} to each block's output:
\begin{equation}
\mathbf{Z}^{(\ell)}
= \mathbf{Z}^{(\ell-1)}
  + \underbrace{\sigma\!\left(\mathbf{W}_g\,\mathbf{V}^{(\ell)}\right)}_{\text{gate}}
    \odot
    \underbrace{\tanh\!\left(\mathbf{W}_f\,\mathbf{V}^{(\ell)}\right)}_{\text{filter}},
\label{eq:gating}
\end{equation}
where $\mathbf{W}_g, \mathbf{W}_f \in \mathbb{R}^{H \times H}$ are learned
projection matrices, $\sigma(\cdot)$ is the sigmoid function, and $\odot$ is
elementwise multiplication.

This mechanism serves as an adaptive information filter. The gate
$\sigma(\mathbf{W}_g\mathbf{V}^{(\ell)}) \in (0,1)^H$ controls how much of the
block's output is incorporated into the running representation:

\begin{itemize}
    \item At \textbf{high noise levels} (large $t$, early diffusion timesteps),
    the signal-to-noise ratio of the block's output is inherently low. The gate
    can suppress noisy or unstable updates, stabilizing the residual accumulation
    across blocks.

    \item At \textbf{low noise levels} (small $t$, fine-detail recovery), the
    gate opens more broadly, allowing the filter $\tanh(\mathbf{W}_f\mathbf{V}^{(\ell)})$
    to refine the representation with stronger feature updates.
\end{itemize}

\noindent The gating mechanism therefore provides an implicit adaptive capacity
that complements the curriculum's explicit difficulty control: even within a given
curriculum stage, individual samples vary in effective difficulty, and the gate
can modulate the update magnitude accordingly.

After $L$ stacked blocks, the final hidden representation
$\mathbf{Z}^{(L)} \in \mathbb{R}^{T \times H}$ is projected by a linear output
layer back to the data dimension $D$, yielding the predicted noise
$\hat{\boldsymbol{\varepsilon}} \in \mathbb{R}^{T \times D}$. Only target-mask
positions contribute to the training loss.

\subsection{Dual-Axis Curriculum Learning}
\label{subsec:curriculum}

In the masked conditional diffusion setting, reconstruction difficulty arises
from two independent and compounding sources:

\begin{enumerate}
    \item \textbf{Masking difficulty.} As the masking ratio $m$ increases, a
    smaller fraction of observed values is available as conditioning context. At
    $m=0.1$, 90\% of observed values remain visible, providing dense temporal and
    cross-variable context; at $m=0.5$, half of the observed values are withheld,
    requiring the model to make broader inferences from sparse context. The
    effective difficulty scales monotonically with $m$.

    \item \textbf{Diffusion difficulty.} At large timesteps $t$ (near $N$), the
    target entries are almost entirely overwritten with Gaussian noise:
    $\mathrm{SNR}(t) = \bar{\alpha}_t/(1-\bar{\alpha}_t) \approx 0$. Denoising
    requires the model to reconstruct the signal almost from scratch, relying
    primarily on the conditioning context and learned prior structure. At small
    $t$ (near 0), the corrupted target values are close to the clean signal, and
    denoising requires precise local refinement under minimal corruption.
\end{enumerate}

Standard diffusion training samples timesteps uniformly from $\mathrm{Uniform}(1,N)$
and applies a fixed masking ratio throughout training. This immediately exposes
the model to the full joint difficulty spectrum including the hardest
combinations of high masking and high noise before the model has acquired even
basic denoising competence. We argue this is the root cause of slow convergence
and large residual variance in standard diffusion-based anomaly detectors.
Our curriculum explicitly manages this difficulty progression.

\subsubsection{Formal Difficulty Measure}
\label{subsubsec:difficulty}

To provide a principled basis for curriculum design, we define a scalar
\emph{reconstruction difficulty measure} that combines both axes. Let
\begin{equation}
\mathrm{SNR}(t)
= \frac{\bar{\alpha}_t}{1-\bar{\alpha}_t}
\label{eq:snr}
\end{equation}
denote the signal-to-noise ratio at diffusion step $t$. The difficulty proxy is:
\begin{equation}
\mathcal{D}(m,t)
= m \cdot \frac{1}{\mathrm{SNR}(t)+\epsilon}
= m \cdot \frac{1-\bar{\alpha}_t}{\bar{\alpha}_t + \epsilon(1-\bar{\alpha}_t)},
\label{eq:difficulty}
\end{equation}
where $\epsilon > 0$ is a small numerical stabilizer. This quantity is:
\begin{itemize}
    \item \textbf{Monotone increasing in $m$}: more withheld observations
    reduces available context, making reconstruction harder.
    \item \textbf{Monotone increasing in $t$}: lower SNR reduces recoverable
    signal in the target entries, demanding more inference under uncertainty.
    \item \textbf{Multiplicative}: the two axes interact heavy masking under
    high noise is substantially harder than either in isolation. This reflects
    the intuition that reconstructing withheld values becomes much harder when
    those values have also been overwritten with noise.
\end{itemize}

Note that $\mathcal{D}(m,t)$ is used as a \emph{theoretical justification}
for the curriculum design rather than as an explicit optimization objective.
It provides a formal basis for ordering training configurations along a
monotone difficulty axis, supporting the choice of the three-stage schedule
below.

\subsubsection{Stage-Wise Curriculum Schedule}
\label{subsubsec:schedule}

We implement the curriculum via a three-stage training schedule
($K_c=3$). At stage $k \in \{1,2,3\}$, both the masking ratio and the
diffusion timestep range are set to curriculum-appropriate values:

\paragraph{Masking schedule.}
The masking ratio is fixed per stage:
\begin{equation}
m_k \in \{0.1,\; 0.3,\; 0.5\},
\quad k \in \{1,2,3\}.
\label{eq:mask_schedule}
\end{equation}
At stage 1, only 10\% of observed entries are withheld, leaving 90\% as
conditioning context a nearly complete view of the time series. At stage 3,
50\% are withheld, requiring significant global inference.

\begin{figure}[t]
    \centering
    \paperfigure[width=0.92\linewidth]{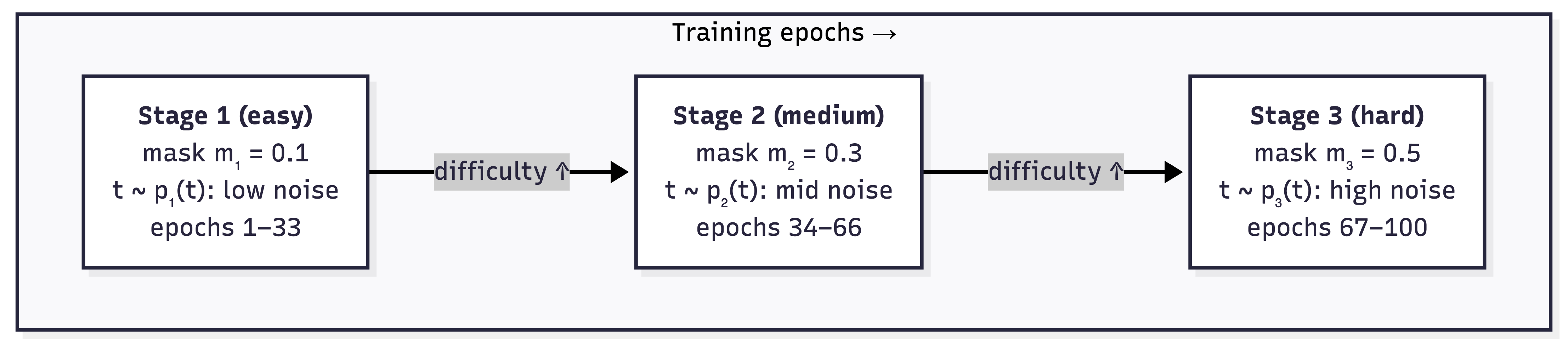}
    \caption{Three-stage dual-axis curriculum schedule. At each stage $k$,
    both the masking ratio $m_k$ and the diffusion timestep distribution
    $p_k(t)$ increase, implementing a monotone easy-to-hard progression
    over the joint difficulty space $\mathcal{D}(m,t)$.
    Each stage occupies $\lfloor E/3 \rfloor$ training epochs;
    stage boundaries are deterministic and require no hyperparameter tuning.}
    \label{fig:curriculum_schedule}
\end{figure}

\paragraph{Timestep schedule.}
Diffusion timesteps are sampled from a stage-specific uniform distribution
over a contiguous sub-range of $[1,N]$:
\begin{equation}
t \sim p_k(t)
= \mathrm{Uniform}\!\left(
    \left\lfloor\tfrac{(k-1)N}{3}\right\rfloor + 1,\;
    \left\lfloor\tfrac{kN}{3}\right\rfloor
  \right),
\quad k \in \{1,2,3\}.
\label{eq:timestep_schedule}
\end{equation}
For $N=1000$, this yields:
\begin{itemize}
    \item \textbf{Stage 1}: $t \sim \mathrm{Uniform}(1, 333)$ low to moderate
    noise; the cosine schedule places these timesteps in the high-SNR regime
    where corrupted targets are still close to the clean signal.
    \item \textbf{Stage 2}: $t \sim \mathrm{Uniform}(334, 667)$ moderate
    noise; the model must handle intermediate corruption where neither local
    refinement nor global reconstruction alone suffices.
    \item \textbf{Stage 3}: $t \sim \mathrm{Uniform}(668, 1000)$ high noise;
    the model must reconstruct heavily corrupted targets from sparse context,
    requiring it to draw heavily on learned temporal priors.
\end{itemize}

\paragraph{Stage duration.}
If the total training budget is $E$ epochs, each stage occupies
$\lfloor E/3 \rfloor$ epochs (with any remainder allocated to the final stage).
This linear partition is deterministic and requires no hyperparameter tuning
for the stage boundaries.

\paragraph{Relationship to the difficulty measure.}
The three-stage design partitions the joint difficulty space
$\mathcal{D}(m,t)$ into three non-overlapping regions, corresponding
to easy ($k=1$), medium ($k=2$), and hard ($k=3$) training regimes.
Within each stage, stochasticity from the uniform timestep distribution
and from random masking assignments provides sufficient diversity to
prevent overfitting to a single difficulty level.

\subsubsection{Optimization Intuition and Theoretical Motivation}
\label{subsubsec:curriculum_theory}

The curriculum is intended to reduce \emph{optimization variance} the
variance of the gradient signal across training iterations in early
training. When the model is initialized with random weights, it cannot yet
perform meaningful denoising. Exposing it immediately to hard tasks (high $m$,
high $t$) produces gradients dominated by noise rather than signal, slowing
convergence. Under the easy curriculum of stage 1, the model can instead
establish stable representations of local temporal structure and
inter-variable correlations by solving well-conditioned reconstruction tasks.
These early representations then serve as a foundation for the harder stages.

Formally, we can interpret the curriculum as sequentially expanding the
\emph{effective reconstruction manifold} that the denoiser must cover.
At stage 1, the manifold is restricted to low-corruption, high-context
settings a compact, well-conditioned subproblem. At stage 3, the manifold
expands to cover the full difficulty range. This progressive expansion
reduces the norm of the initialization-to-solution distance at each stage
transition, which is consistent with classical curriculum learning theory
showing improved convergence guarantees under ordered task sequences
\citep{bengio2009curriculum}.

This is particularly relevant for the anomaly detection application.
The key objective is not just reconstruction accuracy, but the
\emph{separability of residuals}: anomalous inputs should produce higher
residuals than normal inputs. A model that has learned stable, low-variance
reconstructions of normal temporal patterns under structured curricula
is better positioned to produce large, consistent reconstruction errors
on anomalous inputs that deviate from learned temporal priors.

\subsection{Training Objective}
\label{subsec:loss}

\subsubsection{Primary Diffusion Loss}
\label{subsubsec:diff_loss}

The primary learning signal is the masked noise prediction loss, computed
exclusively over target entries:
\begin{equation}
\mathcal{L}_{\mathrm{diff}}
= \mathbb{E}_{t \sim p_k, \boldsymbol{\varepsilon} \sim \mathcal{N}(\mathbf{0},\mathbf{I})}
  \left[
    \left\|
      \mathbf{M}_{\mathrm{tar}}
      \odot
      \bigl(\boldsymbol{\varepsilon}
            - \varepsilon_\theta(\mathbf{X}_t, t, \mathbf{C})\bigr)
    \right\|_2^2
  \right].
\label{eq:diff_loss}
\end{equation}
The expectation is over both the curriculum-stage timestep distribution
$p_k(t)$ and independently sampled Gaussian noise $\boldsymbol{\varepsilon}$.
The $\mathbf{M}_{\mathrm{tar}}$ mask ensures that the loss is computed
only on withheld observed entries: conditioning entries (correctly passed
to the model as context) and structurally missing entries (excluded
from the observation set) do not contribute to the gradient signal.

This objective is equivalent to maximizing the evidence lower bound (ELBO)
of a conditional DDPM restricted to target entries, under the noise
prediction parameterization. The curriculum-stage distribution $p_k(t)$
replaces the standard uniform timestep sampling, implementing the planned
difficulty progression.

\subsubsection{Velocity Consistency Regularization}
\label{subsubsec:vel_loss}

Pure denoising objectives optimize pixel-level fidelity at each target
entry independently. In multivariate time series, however, plausible
reconstructions must also satisfy temporal continuity: the first-order
temporal derivative (velocity) of the reconstructed signal should match
that of the original signal where both entries are observed. Without this
constraint, the denoiser may produce locally accurate but temporally jagged
reconstructions, where adjacent time steps take implausibly large deviations.
Jagged reconstructions inflate reconstruction residuals at non-anomalous
transitions, degrading anomaly score separability.

We define the first-order temporal difference at position $(\tau,d)$ as:
\begin{equation}
\Delta\mathbf{X}_{\tau,d}
= \mathbf{X}_{\tau,d} - \mathbf{X}_{\tau-1,d},
\qquad \tau=2,\dots,T.
\label{eq:first_diff}
\end{equation}
The velocity consistency loss penalizes mismatches between predicted and
true first-order differences, restricted to positions where both consecutive
entries are observed:
\begin{equation}
\mathcal{L}_{\mathrm{vel}}
= \frac{
    \displaystyle\sum_{\tau=2}^{T}\sum_{d=1}^{D}
      \mathbf{M}^{(1)}_{\tau,d}
      \bigl|\Delta\hat{\mathbf{X}}_{0,\tau,d}
            - \Delta\mathbf{X}_{0,\tau,d}\bigr|
  }{
    \displaystyle\sum_{\tau=2}^{T}\sum_{d=1}^{D}\mathbf{M}^{(1)}_{\tau,d}
  },
\label{eq:vel_loss}
\end{equation}
where $\mathbf{M}^{(1)}_{\tau,d} = \mathbf{M}_{\mathrm{obs},\tau,d} \cdot
\mathbf{M}_{\mathrm{obs},\tau-1,d}$ is a joint validity mask that equals 1
only when both $(\tau,d)$ and $(\tau-1,d)$ are observed. This prevents
penalizing derivative mismatches across structurally missing transitions,
where the notion of temporal continuity is undefined. The use of absolute
value (L1 penalty) rather than squared error makes the velocity loss more
robust to occasional large-derivative patterns, such as step changes in
industrial signals.

\subsubsection{Acceleration Consistency Regularization}
\label{subsubsec:acc_loss}

Analogously, we enforce second-order temporal smoothness through an
acceleration consistency term. Define the second-order temporal difference:
\begin{equation}
\Delta^2\mathbf{X}_{\tau,d}
= \mathbf{X}_{\tau,d}
  - 2\mathbf{X}_{\tau-1,d}
  + \mathbf{X}_{\tau-2,d},
\qquad \tau=3,\dots,T.
\label{eq:second_diff}
\end{equation}
The acceleration loss is:
\begin{equation}
\mathcal{L}_{\mathrm{acc}}
= \frac{
    \displaystyle\sum_{\tau=3}^{T}\sum_{d=1}^{D}
      \mathbf{M}^{(2)}_{\tau,d}
      \bigl|\Delta^2\hat{\mathbf{X}}_{0,\tau,d}
            - \Delta^2\mathbf{X}_{0,\tau,d}\bigr|
  }{
    \displaystyle\sum_{\tau=3}^{T}\sum_{d=1}^{D}\mathbf{M}^{(2)}_{\tau,d}
  },
\label{eq:acc_loss}
\end{equation}
where $\mathbf{M}^{(2)}_{\tau,d} = \mathbf{M}_{\mathrm{obs},\tau,d} \cdot
\mathbf{M}_{\mathrm{obs},\tau-1,d} \cdot \mathbf{M}_{\mathrm{obs},\tau-2,d}$
requires three consecutive observed entries. The acceleration term discourages
abrupt changes in the rate of change of the reconstructed signal, enforcing
physical plausibility for smooth-process time series common in manufacturing
and environmental sensing applications. Together, the velocity and acceleration terms encourage the denoiser to produce
reconstructions that lie on a smoother manifold than those achievable through
noise prediction alone. This is analogous to regularization via smoothness
priors commonly used in Gaussian process regression or physically informed
neural networks, adapted here to the diffusion training objective.

\subsubsection{Final Combined Loss and Weighting}
\label{subsubsec:final_loss}

The overall training objective combines the three terms:
\begin{equation}
\mathcal{L}
= \mathcal{L}_{\mathrm{diff}}
  + \lambda_{\mathrm{vel}}\,\mathcal{L}_{\mathrm{vel}}
  + \lambda_{\mathrm{acc}}\,\mathcal{L}_{\mathrm{acc}},
\label{eq:total_loss}
\end{equation}
with weights $\lambda_{\mathrm{vel}}=0.1$ and $\lambda_{\mathrm{acc}}=0.05$.
These values were selected to ensure that the regularization terms provide
a meaningful but non-dominant contribution to the gradient signal: the
primary reconstruction objective $\mathcal{L}_{\mathrm{diff}}$ drives the
core denoising capability, while the derivative terms act as lightweight
regularizers that shape the geometry of the reconstruction manifold.
Setting $\lambda_{\mathrm{acc}} < \lambda_{\mathrm{vel}}$ reflects the
intuition that first-order continuity is a stronger prior than
second-order smoothness for the industrial time series in our benchmark.

\paragraph{Interaction with the curriculum.}
The derivative regularization terms interact beneficially with the
curriculum. At stage 1 (low masking, low noise), the model is reconstructing
small target segments under minimal corruption, and the velocity and
acceleration terms provide precise supervision signals with low ambiguity.
At stage 3 (heavy masking, high noise), derivative supervision is
coarser fewer transitions between consecutive observed target entries are
available but the model has already internalized smooth temporal dynamics
from the earlier stages. This interaction reinforces the easy-to-hard
learning structure across all three loss components.

\subsection{Inference and Anomaly Scoring}
\label{subsec:inference}

\subsubsection{Stochastic Reconstruction via Reverse Diffusion}
\label{subsubsec:reverse_inference}

At inference time, for each test window the model receives the observed entries
$\mathbf{M}_{\mathrm{obs}} \odot \mathbf{X}$ and reconstructs the target entries
through the learned reverse diffusion process. The complementary split-mask
strategy is used: exactly half the observed entries are assigned to
$\mathbf{M}_{\mathrm{cond}}$ and half to $\mathbf{M}_{\mathrm{tar}}$ (alternating
in a deterministic pattern), ensuring full coverage of the observed signal over
two reconstruction passes if needed.

Rather than  DDPM sampling (which requires $N=1000$ reverse steps),
we employ \textbf{DPM-Solver++}\cite{lu2022dpm} with $K=25$ reverse steps.
DPM-Solver++ is a high-order ODE solver specifically designed for diffusion
model sampling that achieves comparable reconstruction quality to DDPM in a
fraction of the steps by exploiting the semi-linear structure of the diffusion
ODE. This reduces per-window inference time by approximately $40\times$ compared
to full DDPM sampling (from $\sim$500ms to $\sim$12ms per window), making
near-real-time anomaly scoring feasible in online monitoring settings.

\paragraph{Monte Carlo ensemble aggregation.}
Because the reverse diffusion process is stochastic different noise
initializations lead to different reconstructions we generate $S=8$ independent
samples per window:
\begin{equation}
\hat{\mathbf{X}}_0^{(1)},\; \hat{\mathbf{X}}_0^{(2)},\; \dots,\; \hat{\mathbf{X}}_0^{(S)}.
\label{eq:samples}
\end{equation}
These are aggregated elementwise using the \emph{median}:
\begin{equation}
\hat{\mathbf{X}}_0
= \mathrm{median}_{s=1,\dots,S}\hat{\mathbf{X}}_0^{(s)}.
\label{eq:median_agg}
\end{equation}
Median aggregation is preferred over the mean for two reasons. First, it is
robust to occasional outlier samples arising from unstable reverse-process
trajectories, which can occur when the model is applied out-of-distribution.
Second, the median estimate has lower asymptotic variance than the mean under
heavy-tailed residual distributions, which are characteristic of anomaly
detection settings where the reconstruction error distribution is highly
skewed.

\subsubsection{Pointwise Anomaly Score Computation}
\label{subsubsec:scoring}

For each time step $\tau$ in the test window, the anomaly score is computed
as the normalized mean absolute reconstruction residual over observed channels:
\begin{equation}
r_\tau
= \frac{
    \sum_{d=1}^{D}
      \mathbf{M}_{\mathrm{obs},\tau,d}
      \left|\mathbf{X}_{\tau,d} - \hat{\mathbf{X}}_{0,\tau,d}\right|
  }{
    \sum_{d=1}^{D}\mathbf{M}_{\mathrm{obs},\tau,d} + \epsilon
  }.
\label{eq:pointwise_score}
\end{equation}
Normalization by the number of observed channels at time $\tau$ is essential
because the observation density varies across time in the presence of structural
missingness: without normalization, time steps with fewer observed channels
would systematically receive lower scores regardless of reconstruction quality,
introducing a spurious correlation between missingness rate and anomaly score.

The score $r_\tau$ can be interpreted as the expected absolute reconstruction
error per observed variable at time $\tau$: under the generative model, normal
inputs should produce small residuals across all channels, while anomalous inputs
should produce large residuals in at least a subset of channels. Crucially, the
curriculum-trained model is expected to produce \emph{lower} residuals on normal
inputs (having learned a tighter generative model of normal dynamics) and
\emph{higher} residuals on anomalous inputs, yielding better score separability.

\subsubsection{Window-Level Score and Threshold Decision}
\label{subsubsec:threshold}

The window-level anomaly score aggregates pointwise scores over the full window:
\begin{equation}
\mathcal{A}
= \frac{1}{T}\sum_{\tau=1}^{T} r_\tau.
\label{eq:window_score}
\end{equation}
This scalar summary is useful for segment-level anomaly reporting but is not
the primary evaluation unit; pointwise scores $r_\tau$ are used for
timestamp-level detection.

Binary anomaly predictions are obtained by thresholding:
\begin{equation}
\hat{y}_\tau = \mathbb{I}(r_\tau \geq \gamma),
\label{eq:decision}
\end{equation}
where $\gamma$ is selected on a held-out validation set by maximizing the F1
score over validation data. \emph{Test data is never used for threshold
calibration}, ensuring that all reported detection performance reflects genuine
generalization. 

\section{Experimental Setup}
\label{sec:experimental_setup}

We evaluate NV-Tesseract-AD under two complementary regimes: large-scale
heterogeneous training on a unified multivariate corpus, and zero-shot
cross-domain transfer to held-out benchmark families without any per-dataset
fine-tuning. This section describes the datasets, evaluation metrics, and
implementation details that underpin both regimes.

\subsection{Datasets}
\label{subsec:datasets}

\paragraph{Training corpus: TSB-UAD-M.}
We train NV-Tesseract-AD on \textbf{TSB-UAD-M}, a multivariate corpus derived
from the TSB-AD benchmark suite~\cite{liu2024elephant}. TSB-AD comprises
1{,}070 labeled time series (870 univariate, 200 multivariate), totaling 55.3M
time points with 1.85M anomalous points (overall anomaly ratio 3.34\%). The
multivariate subset spans 2--248 channels per file across highly heterogeneous
domains, including industrial control telemetry, web-service metrics, physical
sensors, synthetic generators, and human-activity streams.

Several preprocessing steps are required to unify this heterogeneous corpus for
diffusion training. First, all series are standardized to a target dimensionality
of $D_{\text{target}} = 64$: files exceeding this are projected via principal
component analysis (PCA) retaining 95\% of explained variance, while files with
fewer channels are zero-padded to 64 dimensions. The reported raw channel count
for any given evaluation file therefore reflects the original dimensionality
before padding. Second, each channel is independently normalized using robust
statistics with percentile-based clipping at the 1st and 99th percentiles,
followed by scaling to a bounded range; this step was found essential for stable
diffusion training under the large inter-file variance shifts present in
TSB-UAD-M. Finally, because anomalies in TSB-AD are highly concentrated in the
benchmark's original test split, we construct a stratified 70/30 split over the
merged corpus to ensure representative coverage in both partitions. This yields
approximately 37.4M normal and 1.29M anomalous points for training, and 16.0M
normal and 0.55M anomalous points for testing.

\paragraph{Cross-domain evaluation datasets.}
To assess generalization beyond the training distribution, we additionally
evaluate on five external held-out datasets that were not seen during training
and differ substantially in sampling rate, channel count, anomaly definition,
and application domain: \textit{Wafer Manufacturing}~\cite{dau2019ucr},
\textit{CalIt2 Building People Counts}~\cite{hutchins2006calit2},
\textit{Network Traffic}~\cite{kaggle_nettraffic},
\textit{Genesis Demonstrator}~\cite{kaggle_genesis}, and
\textit{Falling People}~\cite{kaggle_falling}. Three of these datasets--Wafer,
Network Traffic, and Falling People--are used in the sampler efficiency
comparison of \S\ref{subsec:sampler_results}; CalIt2 and Genesis appear as
components of the EasyTSAD suite described below. For broader zero-shot
evaluation across standardized benchmark families, we report results on the
\textbf{EasyTSAD} benchmark suite~\cite{easytsad}, which covers ten dataset
families with distinct anomaly semantics and labeling conventions.

\subsection{Evaluation Metrics}
\label{subsec:metrics}

We report anomaly detection metrics, reconstruction fidelity metrics, and a
training-dynamics metric. Let $\mathbf{X} \in \mathbb{R}^{T \times D}$ denote
the observed time series, $\hat{\mathbf{X}}_0$ the reconstructed output, and
$\mathbf{y} \in \{0,1\}^T$ the ground-truth anomaly labels, where $y_\tau = 1$
indicates an anomalous timestamp.

\paragraph{Anomaly scoring.}
For each timestamp $\tau$, the pointwise anomaly score $r_\tau$ is the
normalized mean absolute reconstruction residual over observed channels:
\begin{equation}
r_\tau
=
\frac{\sum_{d=1}^{D} M_{\mathrm{obs},\tau,d}
      \left|X_{\tau,d} - \hat{X}_{0,\tau,d}\right|}
     {\sum_{d=1}^{D} M_{\mathrm{obs},\tau,d} + \epsilon},
\label{eq:anomaly_score}
\end{equation}
where $\mathbf{M}_{\mathrm{obs}}$ is the observation mask and $\epsilon > 0$
is a small numerical stabilizer. Normalizing by the number of observed channels
ensures that timestamps with higher structural missingness do not systematically
receive lower scores. Binary predictions are obtained by thresholding:
\begin{equation}
\hat{y}_\tau = \mathbb{I}(r_\tau \geq \gamma),
\label{eq:threshold}
\end{equation}
where $\gamma$ is selected on held-out validation data by maximizing F1-score.
\emph{Test data is never used for threshold calibration.} In \S\ref{subsec:thresholding_results}
we additionally report two analysis-only reference bounds--a grid-searched SCS
configuration and an oracle AdjBestF1 bound where $\gamma$ is chosen using
ground-truth labels--to quantify headroom above the standard unsupervised
pipeline. These bounds are not part of the deployed evaluation protocol and
are used solely to localize the bottleneck.

\paragraph{Point-adjusted detection metrics.}
We compute standard precision, recall, and F1 under the point-adjusted protocol
used by the EasyTSAD benchmark~\cite{easytsad}:
\begin{equation}
\mathrm{Precision} = \frac{\sum_\tau \hat{y}_\tau y_\tau}{\sum_\tau \hat{y}_\tau},
\quad
\mathrm{Recall} = \frac{\sum_\tau \hat{y}_\tau y_\tau}{\sum_\tau y_\tau},
\quad
\mathrm{F1} = \frac{2 \cdot \mathrm{Precision} \cdot \mathrm{Recall}}{\mathrm{Precision} + \mathrm{Recall}}.
\label{eq:det_metrics}
\end{equation}
Point-adjusted F1 is the primary reported detection metric throughout all
results tables. AUROC is not reported in the main results because the
EasyTSAD evaluation protocol does not define a standard AUROC variant across
its ten families; we retain the scoring formula of Eq.~\eqref{eq:anomaly_score}
for future AUROC-based analysis.

\paragraph{Reconstruction fidelity metrics.}
To assess the quality of imputation independently of threshold selection and
class imbalance, we report masked MAE and MSE over target entries:
\begin{equation}
\mathrm{MAE}
=
\frac{\sum_{\tau,d} M_{\mathrm{tar},\tau,d}\left|X_{\tau,d} - \hat{X}_{0,\tau,d}\right|}
     {\sum_{\tau,d} M_{\mathrm{tar},\tau,d}},
\quad
\mathrm{MSE}
=
\frac{\sum_{\tau,d} M_{\mathrm{tar},\tau,d}\left(X_{\tau,d} - \hat{X}_{0,\tau,d}\right)^2}
     {\sum_{\tau,d} M_{\mathrm{tar},\tau,d}}.
\label{eq:recon_metrics}
\end{equation}
These are reported in the curriculum comparison of \S\ref{subsec:curriculum_effect}
to characterize reconstruction improvement independently of thresholding.

\paragraph{Training reconstruction loss.}
The diffusion noise-prediction loss $\mathcal{L}_{\mathrm{diff}}$ (defined in
\S\ref{subsubsec:diff_loss}) is tracked throughout training to characterize
reconstruction stability under complementary masking strategies. This training-phase
metric is distinct from the downstream MAE and MSE reported at inference time.

\subsection{Implementation Details}
\label{subsec:implementation}

\paragraph{Framework and training infrastructure.}
NV-Tesseract-AD is implemented in PyTorch and trained on NVIDIA A100 GPUs.
Mixed-precision training (FP16) is used throughout to improve computational
throughput while maintaining numerical stability in the diffusion forward and
reverse passes.

\paragraph{Diffusion configuration.}
The model uses a discrete-time DDPM formulation~\cite{ho2020ddpm} with a cosine
variance schedule~\cite{nichol2021improved} over $N = 1{,}000$ diffusion steps.
During training, timesteps are sampled from the curriculum-stage distribution
$p_k(t)$ rather than the standard uniform distribution (see
\S\ref{subsubsec:schedule}). At inference time, we replace ancestral DDPM
sampling with \textbf{DPM-Solver++}~\cite{lu2022dpm} using 25 reverse steps.
This reduces per-window sampling latency from approximately 500\,ms (DDPM,
$N{=}1000$ steps) to approximately 12\,ms (DPM-Solver++, 25 steps)--a
per-window reduction of roughly $40{\times}$. End-to-end dataset-level runtime
reductions are lower ($3.5$--$5.7{\times}$, reported in
\S\ref{subsec:sampler_results}) due to I/O overhead and parallelism effects.
To mitigate stochastic variance across reverse-process trajectories, we generate
$S = 8$ independent samples per window and aggregate using the elementwise
median, which is more robust to occasional unstable samples than the mean.

\paragraph{Inference-time curriculum stage.}
The model's Transformer denoiser receives a curriculum-stage embedding at every
forward pass (see \S\ref{subsubsec:tokens}). At inference time, the stage index
is fixed to $k = 3$, directing the model to apply its most general denoising
policy--the one trained under the heaviest masking and highest noise regime.

\paragraph{Model architecture.}
The diffusion denoiser is a residual Transformer consisting of $L = 8$ stacked
Pre-LayerNorm (Pre-LN) blocks~\cite{xiong2020layer} with hidden dimension
$H = 256$ and 8-head self-attention. Each block applies a feed-forward network
with a $4\times$ expansion ratio and GELU activation, followed by a gated
residual update (sigmoid gate $\otimes$ tanh filter) that stabilizes gradient
flow under high-noise corruption. Diffusion timestep embeddings are generated
via sinusoidal encodings passed through two learned projection layers and added
to all token representations. A summary of all architectural and training
hyperparameters is provided in Table~\ref{tab:hyperparams}.

\begin{table}[t]
\centering
\caption{Key hyperparameters of NV-Tesseract-AD.}
\label{tab:hyperparams}
\begin{tabular}{lc}
\toprule
\textbf{Hyperparameter} & \textbf{Value} \\
\midrule
Transformer blocks $L$                     & 8 \\
Hidden dimension $H$                       & 256 \\
Attention heads                            & 8 \\
FFN expansion ratio                        & $4\times$ \\
Diffusion steps $N$                        & 1{,}000 \\
Variance schedule                          & Cosine \\
Inference steps (DPM-Solver++)             & 25 \\
Inference stage embedding $k$              & 3 \\
Ensemble size $S$                          & 8 \\
Target dimensionality $D_{\text{target}}$  & 64 \\
Window length $L_w$                        & 100 \\
Training stride                            & 10 \\
Evaluation stride                          & 1 \\
Curriculum stages $K_c$                    & 3 \\
Masking ratios $m_k$                       & $\{0.1, 0.3, 0.5\}$ \\
\bottomrule
\end{tabular}
\end{table}

\paragraph{Windowing strategy.}
Time series are segmented into sliding windows of length $L_w = 100$ time steps.
During training, a stride of 10 is used to generate sufficient data diversity
while controlling computational cost. During evaluation, the stride is reduced
to 1 to provide fine-grained per-timestep anomaly localization.

\paragraph{Curriculum schedule.}
Training proceeds in three deterministic stages of equal length
($\lfloor E/3 \rfloor$ epochs each, where $E = 100$ total epochs). At stage
$k$, the masking ratio is fixed to $m_k \in \{0.1, 0.3, 0.5\}$ and diffusion
timesteps are sampled uniformly from the $k$-th third of $[1, N]$, implementing
an explicit easy-to-hard schedule over the joint difficulty space (see
\S\ref{subsec:curriculum}).

\paragraph{Optimization.}
The model is optimized with AdamW~\cite{loshchilov2019decoupled} using learning
rate $\eta = 1 \times 10^{-4}$ and weight decay $\lambda = 1 \times 10^{-4}$.
A cosine learning rate schedule with linear warmup over the first 5\% of training
steps prevents instability during early curriculum stages. Gradient norms are
clipped at 1.0 throughout training. Early stopping is applied based on validation
F1-score, and the best validation checkpoint is used for all reported evaluations.

\paragraph{Hardware and runtime.}
Full training on TSB-UAD-M requires approximately 18 hours on a single A100 GPU.
At inference time, DPM-Solver++ with 25 steps yields a per-window latency of
approximately 12\,ms, enabling real-time deployment in streaming monitoring
pipelines. Full-suite EasyTSAD evaluation under DDPM requires approximately 31
hours; under DPM-Solver++ this is reduced to approximately 6--9 hours depending
on parallelism configuration.

\section{Results and Discussion}
\label{sec:results}

We evaluate NV-Tesseract-AD from the perspective that matters most for
reconstruction-based anomaly detection: whether the model produces stable
imputations under structured masking, whether those residuals translate into
useful anomaly scores under unsupervised thresholding, and whether the resulting
detector remains practical at benchmark scale. Since anomaly decisions in our
framework are derived from reconstruction residuals, we first analyze
reconstruction behavior and then study thresholding, solver efficiency, and
cross-domain generalization. Together, these analyses address four central
questions: (1) Does the curriculum stabilize reconstruction and reduce masking
sensitivity? (2) Does curriculum training yield measurable downstream detection
gains? (3) Where does the primary detection bottleneck lie once reconstruction
is stabilized? (4) Does the trained model transfer competitively across unseen
benchmark families, and what are the practical runtime implications of
large-scale deployment?

\subsection{Reconstruction Stability Under Complementary Masking}
\label{subsec:recon_stability}

A fundamental requirement for residual-based anomaly detection is that
reconstruction quality should be robust to the specific masking pattern applied
at inference time. If the model is sensitive to which segments are withheld,
residual scores will be noisy across evaluation runs and threshold-dependent
metrics will be unreliable. Our first question is therefore whether the
curriculum-trained diffusion model learns a reconstruction function that is
genuinely masking-invariant.

To test this, we evaluate checkpoints under two complementary split-mask
strategies, $S_0$ and $S_1$, and track both the average diffusion noise-prediction
loss $\mathcal{L}_{\mathrm{diff}}$ and the discrepancy between them. Let
$\mathcal{L}_0(e)$ and $\mathcal{L}_1(e)$ denote this training-phase
reconstruction loss at epoch $e$ under $S_0$ and $S_1$, respectively, and
define the loss gap:
\begin{equation}
\Delta(e) = |\mathcal{L}_0(e) - \mathcal{L}_1(e)|.
\label{eq:loss_gap}
\end{equation}

\begin{figure}[H]
    \centering
    \paperfigure[width=0.88\linewidth]{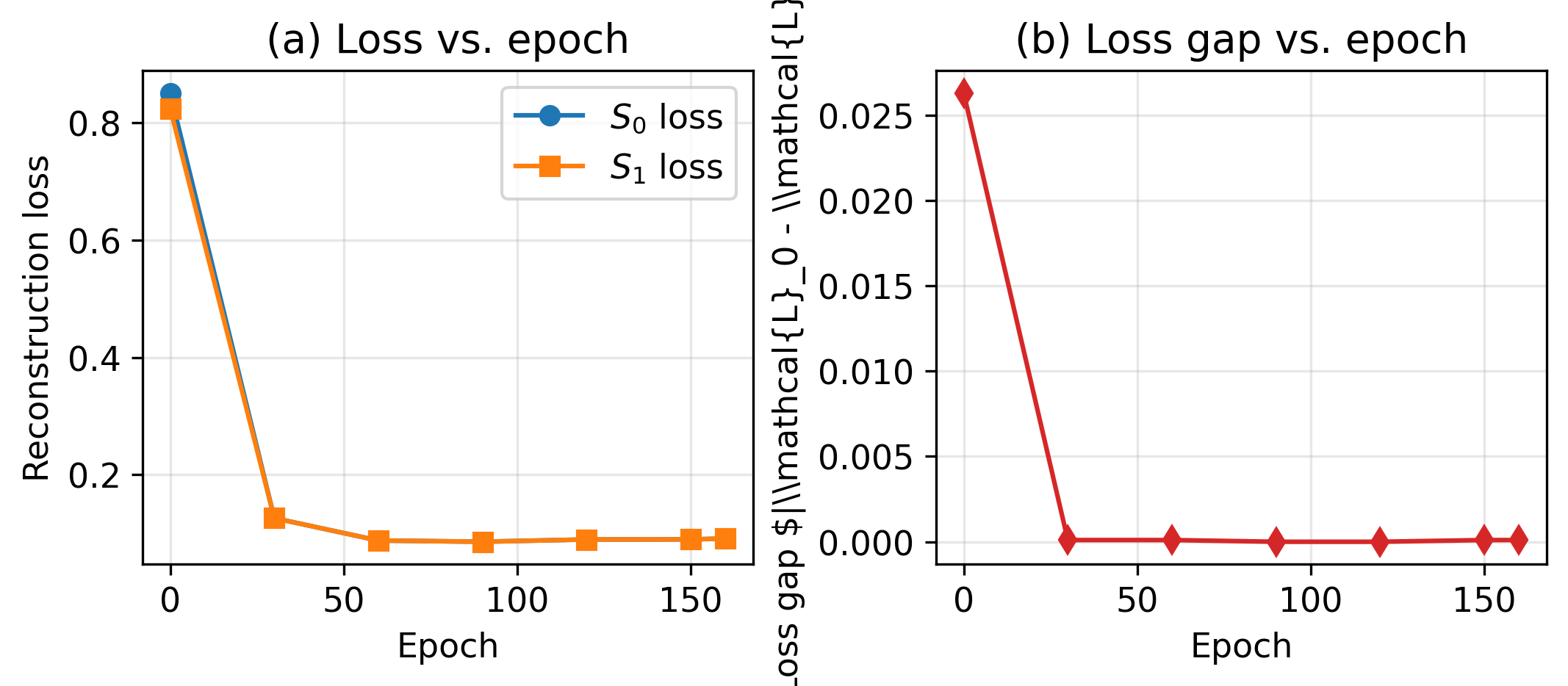}
    \caption{Reconstruction behavior of NV-Tesseract-AD under split-mask
    strategies $S_0$ and $S_1$ over the curriculum training schedule.
    Panel~(a) shows the diffusion noise-prediction loss across epochs; loss
    decreases sharply in early training and stabilizes after epoch~60, with
    the minimum near epoch~90. Panel~(b) shows the loss gap
    $|\mathcal{L}_0 - \mathcal{L}_1|$, which falls from 0.0263 at epoch~0 to
    near zero by epoch~30 and remains negligible thereafter, confirming that
    the curriculum produces masking-invariant reconstructions.}
    \label{fig:recon_loss_gap}
\end{figure}

Figure~\ref{fig:recon_loss_gap} and Table~\ref{tab:recon_loss} reveal a
consistent and interpretable pattern. Reconstruction loss drops sharply during
early training, stabilizes after approximately epoch 60, and reaches its minimum
near epoch 90 with $\mathcal{L}_0 = 0.0856$, $\mathcal{L}_1 = 0.0855$, and
effectively zero gap. Crucially, the loss gap $\Delta(e)$ collapses from 0.0263
at initialization to approximately zero by epoch 30 and remains negligible
throughout the remainder of training. The slight deterioration at epochs
120--160 ($\mathcal{L} \approx 0.0894$--$0.0910$) is consistent with mild
overfitting to the hardest curriculum stage and motivates the early-stopping
criterion based on validation F1.

\begin{table}[]
    \centering
    \caption{Diffusion noise-prediction loss $\mathcal{L}_{\mathrm{diff}}$ and
    loss gap across training epochs for complementary split-mask strategies
    $S_0$ and $S_1$. The loss gap collapses rapidly within the first curriculum
    stage and remains negligible thereafter, indicating robustness to masking
    variation. Epoch~90 yields the lowest loss and zero gap and is selected as
    the evaluation checkpoint.}
    \label{tab:recon_loss}
    \begin{tabularx}{\textwidth}{CCCC}
        \toprule
        \textbf{Epoch} & \textbf{$S_0$ Avg.\ Loss} & \textbf{$S_1$ Avg.\ Loss} & \textbf{Loss Gap $\Delta(e)$} \\
        \midrule
        0   & 0.8511 & 0.8248 & 0.0263 \\
        30  & 0.1258 & 0.1258 & 0.0001 \\
        60  & 0.0876 & 0.0876 & 0.0001 \\
        90  & 0.0856 & 0.0855 & 0.0000 \\
        120 & 0.0894 & 0.0894 & 0.0000 \\
        150 & 0.0895 & 0.0895 & 0.0001 \\
        160 & 0.0910 & 0.0910 & 0.0001 \\
        \bottomrule
    \end{tabularx}
\end{table}

This result carries two practical implications. First, the curriculum does not
merely reduce average reconstruction error; it also eliminates sensitivity to
the specific structured mask applied at test time. This is precisely the
stability property required for reliable residual-based anomaly detection: a
model that reconstructs consistently regardless of masking pattern will produce
stable, threshold-amenable residual distributions. Second, the rapid collapse
of $\Delta(e)$ within the first 30 epochs--corresponding entirely to the first
curriculum stage--suggests that the easy low-masking, low-noise tasks are
sufficient to establish the coarse temporal structure needed for masking
invariance, before the model is exposed to harder settings in stages 2 and 3.
This provides empirical support for the theoretical motivation of the
easy-to-hard curriculum design: early learning under well-conditioned tasks
builds the representational foundation that generalizes across inference-time
masking regimes. Finally, the near-zero loss gap provides a principled basis
for checkpoint selection: epoch 90, which achieves the best reconstruction
fidelity and the strongest masking invariance simultaneously, is used as the
primary evaluation checkpoint throughout all subsequent experiments.

\subsection{Effect of Curriculum Training and Train--Inference Alignment}
\label{subsec:curriculum_effect}

Having established that the curriculum stabilizes reconstruction, we next isolate
the practical impact of curriculum training on downstream anomaly detection
performance. Table~\ref{tab:weights_comparison} compares an earlier model trained
without staged masking or timestep scheduling (2025 baseline) against
NV-Tesseract-AD (curriculum-trained, 2026 weights) under a common evaluation
pipeline on the TSB-UAD-M test split (5{,}000 rows; 38 raw input features,
zero-padded to $D_{\text{target}}=64$ for model input; anomaly rate 1.18\%).
Both models share the identical Transformer diffusion backbone; the only
differences are the training procedure and the consistency of preprocessing
between training and inference.

\begin{table}[H]
    \centering
    \caption{Baseline (non-curriculum, 2025 weights) versus NV-Tesseract-AD
    (curriculum-trained, 2026 weights) under the same inference pipeline and
    evaluation set. Both models share the same Transformer diffusion backbone.
    MAE and MSE are computed over target entries at inference time and are
    distinct from the training-phase noise-prediction loss reported in
    \S\ref{subsec:recon_stability}.}
    \label{tab:weights_comparison}
    \begin{tabularx}{\textwidth}{CCCC}
        \toprule
        \textbf{Metric} & \textbf{Baseline (2025)} & \textbf{NV-Tesseract-AD (2026)} & \textbf{Relative Change} \\
        \midrule
        F1        & 0.0376 & 0.1404 & $\approx 3.7\times$ \\
        Precision & 0.0571 & 0.1882 & $\approx 3.3\times$ \\
        Recall    & 0.0280 & 0.1119 & $\approx 4.0\times$ \\
        MAE       & 1.90   & 1.25   & $\approx 34\%$ lower \\
        MSE       & 3.66   & 1.62   & $\approx 56\%$ lower \\
        \bottomrule
    \end{tabularx}
\end{table}

The improvements are substantial across all metrics. F1 improves approximately
$3.7\times$, precision $3.3\times$, and recall $4.0\times$, while MAE drops
34\% and MSE drops 56\%. Gains of this magnitude cannot be attributed to
incidental variance; they reflect two complementary and separable sources of
improvement. The first is curriculum scheduling itself: staged exposure to
increasing reconstruction difficulty stabilizes gradient dynamics throughout
training, reduces residual variance (as confirmed by the training-loss analysis
in \S\ref{subsec:recon_stability}), and yields a denoiser that reconstructs
normal temporal patterns more tightly, as measured by lower MAE and MSE at
inference. The second is train--inference consistency: matching normalization
statistics, percentile clipping boundaries, masking configuration, inference
schedule, and checkpoint selection between training and evaluation prevents score
distribution mismatch from systematically inflating reconstruction error at test
time.

These two factors are difficult to disentangle in isolation, but their joint
effect on detection performance is clear and large. Critically, this result
demonstrates that in diffusion-based anomaly detection, engineering
consistency--normalization, clipping, masking configuration, inference schedule,
and checkpoint selection--is part of the method rather than a peripheral
implementation detail. A small mismatch in any of these dimensions shifts the
residual score distribution, and once that distribution shifts, unsupervised
thresholding becomes unreliable. Train--inference parity is therefore not merely
good practice; it is a precondition for meaningful evaluation of residual-based
detectors.

\subsection{From Reconstruction Residuals to Anomaly Decisions}
\label{subsec:thresholding_results}

Stable reconstruction alone does not guarantee good anomaly detection, because
a second decision layer is required to convert continuous residual scores $r_\tau$
into binary predictions. We evaluate how much of the final detection quality is
determined by the diffusion model versus by the thresholding mechanism, and where
the primary bottleneck lies.

Residual scores are converted to binary anomaly predictions via unsupervised SCS
and MACS thresholding~\cite{li2026scs-macs}, evaluated on a balanced stress-test
set (5{,}000 anomalous and 45{,}000 normal points from the TSB-UAD-M test split;
anomaly rate $\approx 1.18\%$). Under default SCS, the system achieves
Precision/Recall/F1 of approximately $0.24/0.12/0.16$; default MACS yields
approximately $0.26/0.08/0.12$. These values reflect the conservative bias
typical of unsupervised thresholding under severe class imbalance: the residual
signal is informative, but the thresholder suppresses false positives at the
cost of recall.

To quantify how much headroom remains above default unsupervised performance,
we report two \emph{analysis-only} reference bounds that are not part of the
standard evaluation pipeline and are used solely to localize the bottleneck. A
grid search over 640 SCS hyperparameter combinations identifies a best
configuration--APCA segmentation, confidence level $c = 0.90$, percentile
filter 15, $n_{\mathrm{seg}} = 2$, window factor 3--achieving F1 $\approx
0.218$ on the labeled evaluation slice. An oracle-style AdjBestF1 bound, where
the decision threshold $\gamma$ is selected using ground-truth test labels
rather than held-out validation data, achieves F1 $\approx 0.244$ with
Precision $\approx 0.178$ and Recall $\approx 0.387$. This oracle bound
represents the maximum achievable F1 for any fixed threshold applied to the
current score distribution, and is reported solely as a diagnostic upper limit;
it is never used in the deployed evaluation protocol.

\begin{figure}[H]
    \centering
    \paperfigure[width=0.82\linewidth]{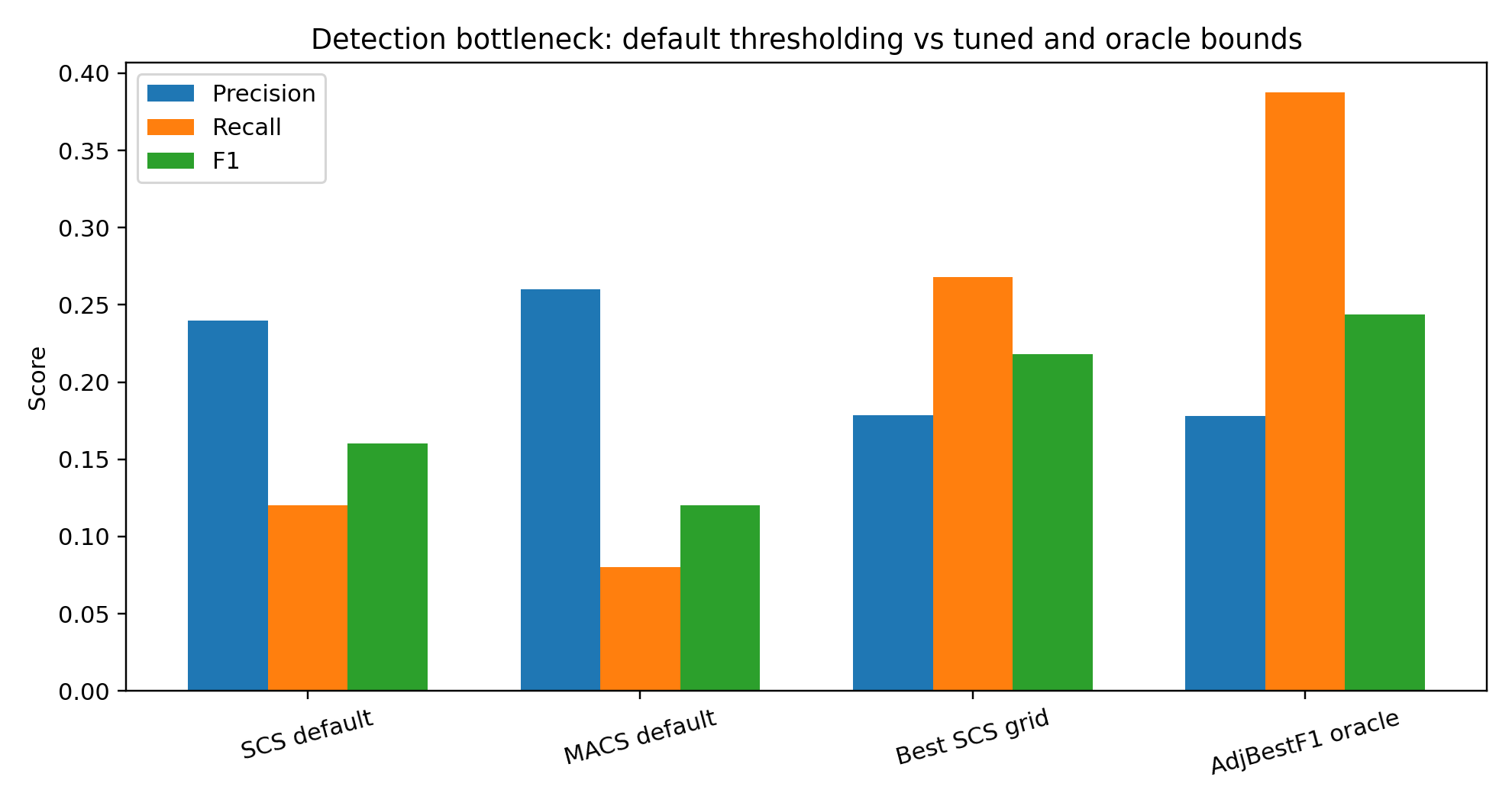}
    \caption{Comparison of default unsupervised thresholding with tuned and
    oracle reference bounds. Bars show precision, recall, and F1 for default
    SCS, default MACS, the best SCS grid-search configuration, and the oracle
    AdjBestF1 upper bound. The oracle is an analysis-only diagnostic computed
    using ground-truth test labels and is not part of the deployed evaluation
    protocol. The gap between default unsupervised thresholding and the oracle
    is substantially larger than the gap between the tuned configuration and
    the oracle, indicating that threshold calibration remains a major bottleneck
    after reconstruction has been stabilized.}
    \label{fig:thresholding_comparison}
\end{figure}

\begin{figure}[H]
    \centering
    \paperfigure[width=0.82\linewidth]{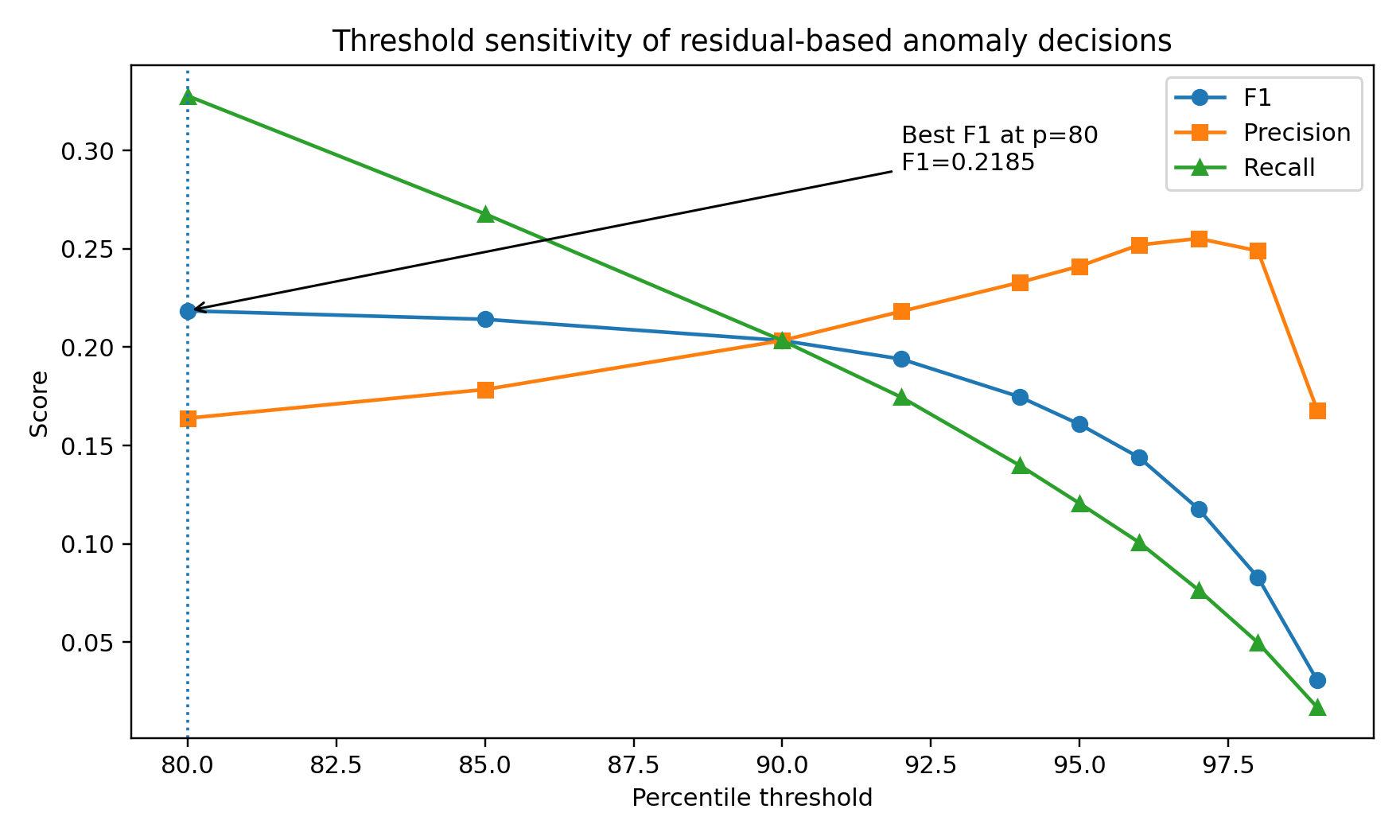}
    \caption{Precision, recall, and F1 as a function of percentile threshold
    on the residual score distribution. The best empirical trade-off appears
    in the 80th--90th percentile region, with peak F1 obtained near the 80th
    percentile. This sensitivity analysis shows that final anomaly detection
    quality depends strongly on calibration of the residual-score threshold
    even when the reconstruction model itself is stable.}
    \label{fig:percentile_sweep}
\end{figure}

The small gap between the grid-search bound (0.218) and the oracle (0.244)
confirms that the reconstruction signal itself is informative: the denoiser is
not the primary bottleneck. The substantially larger gap between default
unsupervised thresholding and the oracle, however, establishes threshold
calibration as the dominant remaining challenge. This is further supported by
the percentile sweep in Figure~\ref{fig:percentile_sweep}: the best empirical
F1 trade-off occurs around the 80th--90th percentile of the residual score
distribution, consistent with the grid-search result, but default unsupervised
thresholds do not automatically locate this region. The implication is
significant: improving the denoiser further will yield diminishing returns until
unsupervised threshold calibration is correspondingly improved. Future work on
score normalization, online calibration under distribution shift, and
anomaly-rate-aware thresholding strategies is therefore likely to yield larger
detection gains than further scaling of the diffusion backbone alone.

\subsection{Zero-Shot Transfer Across EasyTSAD Families}
\label{subsec:easytsad_results}

\begin{figure}[]
    \centering
    \paperfigure[width=0.72\linewidth]{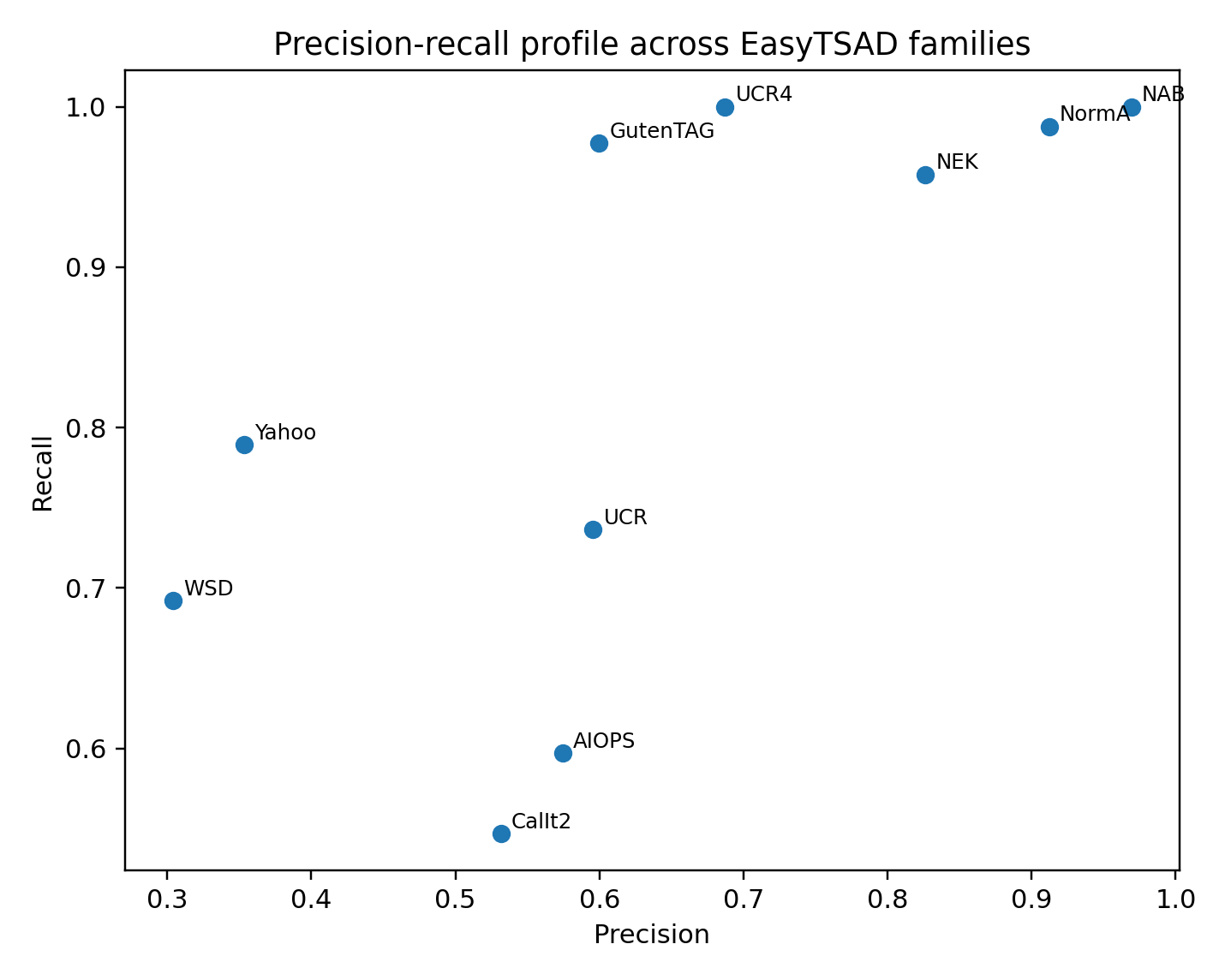}
    \caption{Precision--recall profile across EasyTSAD dataset families.
    Families such as NAB, NormA, and NEK lie in the high-precision/high-recall
    region, whereas Yahoo and WSD exhibit comparatively lower precision despite
    moderate-to-high recall. This pattern is consistent with the interpretation
    that cross-domain performance is limited more by calibration and
    family-specific heterogeneity than by reconstruction quality alone.}
    \label{fig:easytsad_precision_recall}
\end{figure}

We test whether a single NV-Tesseract-AD model trained once on TSB-UAD-M
transfers to unseen benchmark families without dataset-specific tuning. The
model and inference pipeline are applied unchanged to each dataset family in
the EasyTSAD benchmark suite~\cite{easytsad}; anomaly decisions are produced
using fixed SCS/MACS thresholding. Table~\ref{tab:easytsad_f1} reports
point-adjusted Precision, Recall, and F1 by dataset family under the EasyTSAD
evaluation protocol, and Figure~\ref{fig:easytsad_precision_recall} visualizes
the precision--recall profile across families.

\begin{table}[]
    \centering
    \caption{Zero-shot anomaly detection performance of NV-Tesseract-AD on
    EasyTSAD dataset families. The model is trained once on TSB-UAD-M and
    evaluated without per-dataset tuning. Point-adjusted F1 is computed using
    the EasyTSAD evaluation protocol~\cite{easytsad}.}
    \label{tab:easytsad_f1}
    \begin{tabularx}{\textwidth}{CCCC}
        \toprule
        \textbf{Dataset Family} & \textbf{F1} & \textbf{Precision} & \textbf{Recall} \\
        \midrule
        NAB      & 0.9841 & 0.9692 & 1.0000 \\
        NormA    & 0.9477 & 0.9123 & 0.9877 \\
        NEK      & 0.8830 & 0.8261 & 0.9578 \\
        UCR4     & 0.7831 & 0.6872 & 1.0000 \\
        GutenTAG & 0.6928 & 0.5997 & 0.9773 \\
        UCR      & 0.6409 & 0.5957 & 0.7364 \\
        AIOPS    & 0.5557 & 0.5745 & 0.5970 \\
        CalIt2   & 0.5381 & 0.5315 & 0.5468 \\
        Yahoo    & 0.4328 & 0.3536 & 0.7893 \\
        WSD      & 0.3406 & 0.3039 & 0.6923 \\
        \bottomrule
    \end{tabularx}
\end{table}

Performance is strongest on NAB (F1 $= 0.9841$), NormA ($0.9477$), and NEK
($0.8830$), and remains strong on UCR4 ($0.7831$) and GutenTAG ($0.6928$).
Results are more moderate on UCR ($0.6409$), AIOPS ($0.5557$), and CalIt2
($0.5381$). The most challenging families are Yahoo ($0.4328$) and WSD
($0.3406$). The family-wise variation is interpretable and consistent across
all experiments. Families where NV-Tesseract-AD performs strongest share two
characteristics: temporally coherent anomaly semantics and relatively stable
background dynamics, so that reconstruction residuals provide a clean,
well-separated signal that a fixed threshold can reliably exploit. In contrast,
the weaker families--Yahoo and WSD--exhibit high recall (0.69--0.79) but
noticeably lower precision, indicating that the model correctly identifies
suspicious temporal regions but struggles to isolate them cleanly under a
fixed unsupervised threshold. This pattern is consistent with families where
anomaly types, background regimes, and labeling conventions vary strongly
across files, leading to diffuse residual distributions that are difficult
to partition with a single global threshold.

Taken together, the zero-shot results support two simultaneous conclusions.
First, reconstruction-centric diffusion training generalizes across diverse
anomaly detection domains without per-dataset fine-tuning, as evidenced by
strong performance on the majority of EasyTSAD families. This validates the
core design principle of NV-Tesseract-AD: learning a tight generative model of
normal temporal dynamics provides a portable anomaly detection signal. Second,
cross-domain decision quality is still limited by threshold calibration and
family-specific heterogeneity rather than by reconstruction quality alone,
consistent with the thresholding analysis of \S\ref{subsec:thresholding_results}.
Datasets where the reconstruction signal is strong but precision is low suggest
that domain-adaptive calibration--rather than domain-specific retraining--is
the most promising direction for improving zero-shot performance on challenging
families.

\subsection{Fast Samplers Improve Feasibility but Shift Operating Points}
\label{subsec:sampler_results}

\begin{figure}[]
    \centering
    \paperfigure[width=\linewidth]{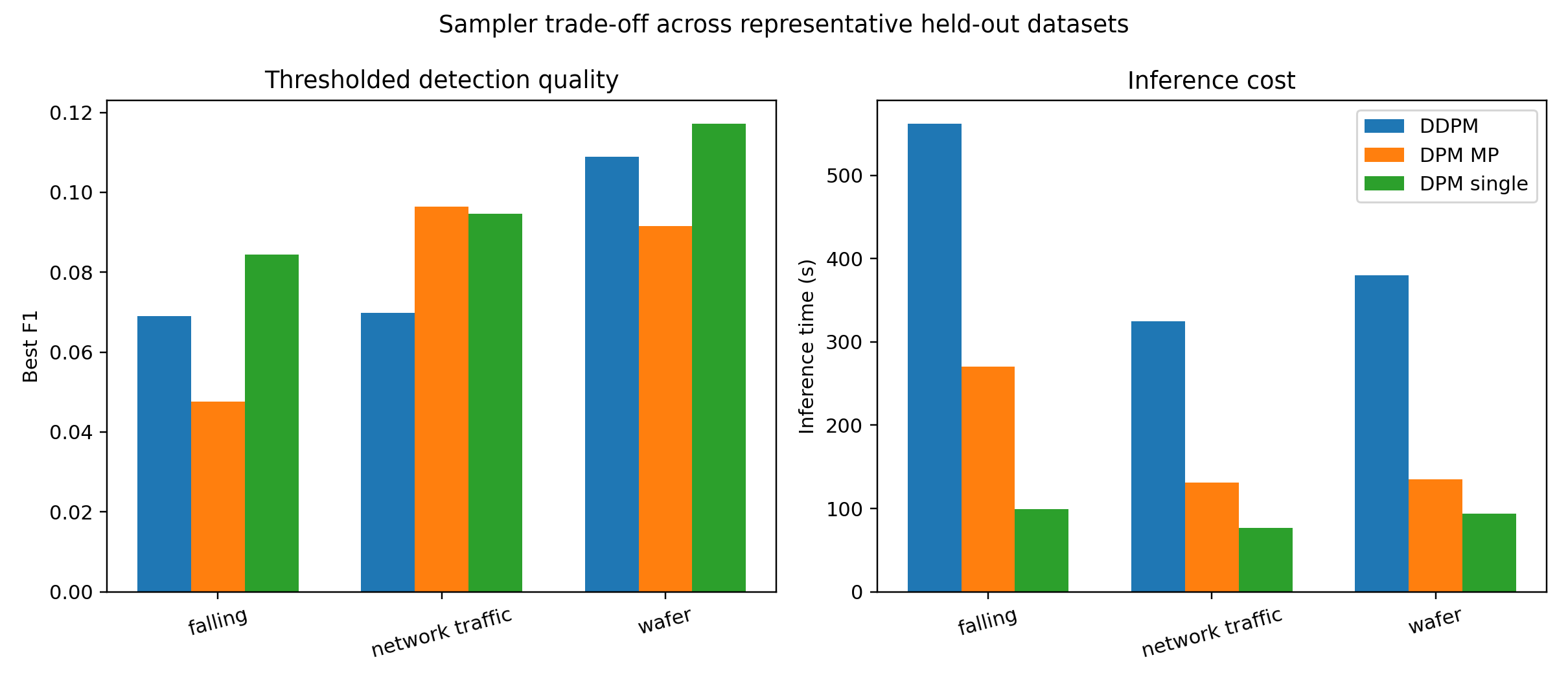}
    \caption{Trade-off between detection quality and runtime for DDPM and
    DPM-Solver++ variants on representative held-out datasets. The left panel
    shows best thresholded F1 under unsupervised calibration, while the right
    panel shows inference time. Faster samplers substantially reduce runtime,
    but also shift the operating point of the detector, demonstrating that
    sampling and thresholding must be validated jointly rather than treated as
    independent design choices.}
    \label{fig:sampler_tradeoff}
\end{figure}

Diffusion-based detectors are often constrained less by modeling capacity than
by sampling cost. Standard DDPM inference requires $N = 1{,}000$ reverse-process
steps, which at approximately 500\,ms per window makes large-scale benchmark
evaluation and real-time deployment prohibitively expensive. We compare DDPM
against DPM-Solver++~\cite{lu2022dpm} (multi-process and single-process
variants) on three of the five held-out datasets introduced in
\S\ref{subsec:datasets}--Wafer Manufacturing, Network Traffic, and Falling
People--reporting oracle F1 upper bounds (UB), best achieved F1 under
unsupervised SCS/MACS thresholding, and wall-clock inference time.
Figure~\ref{fig:sampler_tradeoff} shows the quality--runtime trade-off, and
Table~\ref{tab:ddpm_vs_dpm} provides detailed numerical results.

\begin{table}[]
\centering
\caption{DDPM vs.\ DPM-Solver++ on representative held-out datasets.
DPM~MP: multi-process DPM-Solver++. DPM~S: single-process DPM-Solver++.
UB: oracle best-F1 using ground-truth labels for threshold selection
(analysis-only diagnostic; not the deployed evaluation protocol).
Inference times are dataset-level wall-clock durations; per-window latency
reduction is approximately $40{\times}$ (500\,ms $\to$ 12\,ms), with
lower dataset-level ratios reflecting I/O and parallelism effects.}
\label{tab:ddpm_vs_dpm}
\begin{tabularx}{\textwidth}{lCCCCCCCCC}
\toprule
& \multicolumn{3}{c}{\textbf{Oracle UB (F1)}}
& \multicolumn{3}{c}{\textbf{Best Thresholded F1}}
& \multicolumn{3}{c}{\textbf{Inference Time (s)}} \\
\cmidrule(lr){2-4}\cmidrule(lr){5-7}\cmidrule(lr){8-10}
\textbf{Dataset}
& \textbf{DDPM} & \textbf{DPM MP} & \textbf{DPM S}
& \textbf{DDPM} & \textbf{DPM MP} & \textbf{DPM S}
& \textbf{DDPM} & \textbf{DPM MP} & \textbf{DPM S} \\
\midrule
Falling (2k)    & 0.1043 & 0.1095 & 0.1183 & 0.0690 & 0.0476 & 0.0844 & 562 & 270 & 99 \\
Network Traffic & 0.1818 & 0.1961 & 0.1912 & 0.0698 & 0.0964 & 0.0947 & 325 & 131 & 77 \\
Wafer           & 0.1667 & 0.1798 & 0.2169 & 0.1089 & 0.0916 & 0.1172 & 380 & 135 & 94 \\
\bottomrule
\end{tabularx}
\end{table}

DPM-Solver++ consistently reduces dataset-level inference time by approximately
$3.5$--$5.7\times$ while preserving or slightly improving oracle upper bounds
across all three datasets. This confirms that accelerated sampling does not
degrade the model's underlying reconstruction quality. The effect on thresholded
detection performance is, however, more nuanced. On Network Traffic,
multi-process DPM-Solver++ improves best thresholded F1 from 0.0698 to 0.0964.
On Wafer, multi-process DPM-Solver++ reduces it from 0.1089 to 0.0916, whereas
single-process DPM-Solver++ slightly improves it to 0.1172. A similar mixed
pattern appears across oracle bounds.

This mixed pattern arises because different numbers of reverse steps produce
slightly different residual score distributions, which in turn shift the
effective operating point of fixed SCS/MACS thresholds. The result is an
important systems-level finding: faster samplers are not merely a drop-in
runtime optimization; they alter the reconstruction residual distribution and
therefore the behavior of the downstream thresholding mechanism. Sampler choice
and thresholding cannot be treated as independent design decisions. The correct
interpretation is that DPM-Solver++ makes diffusion-based anomaly detection
feasible at scale, but the resulting detector must be re-validated with the
chosen solver in the loop before deployment--the solver configuration is part
of the model specification, not a transparent implementation detail.

\subsection{Benchmark-Scale Runtime and Practical Implications}
\label{subsec:runtime_results}

\begin{figure}[H]
    \centering
    \paperfigure[width=0.82\linewidth]{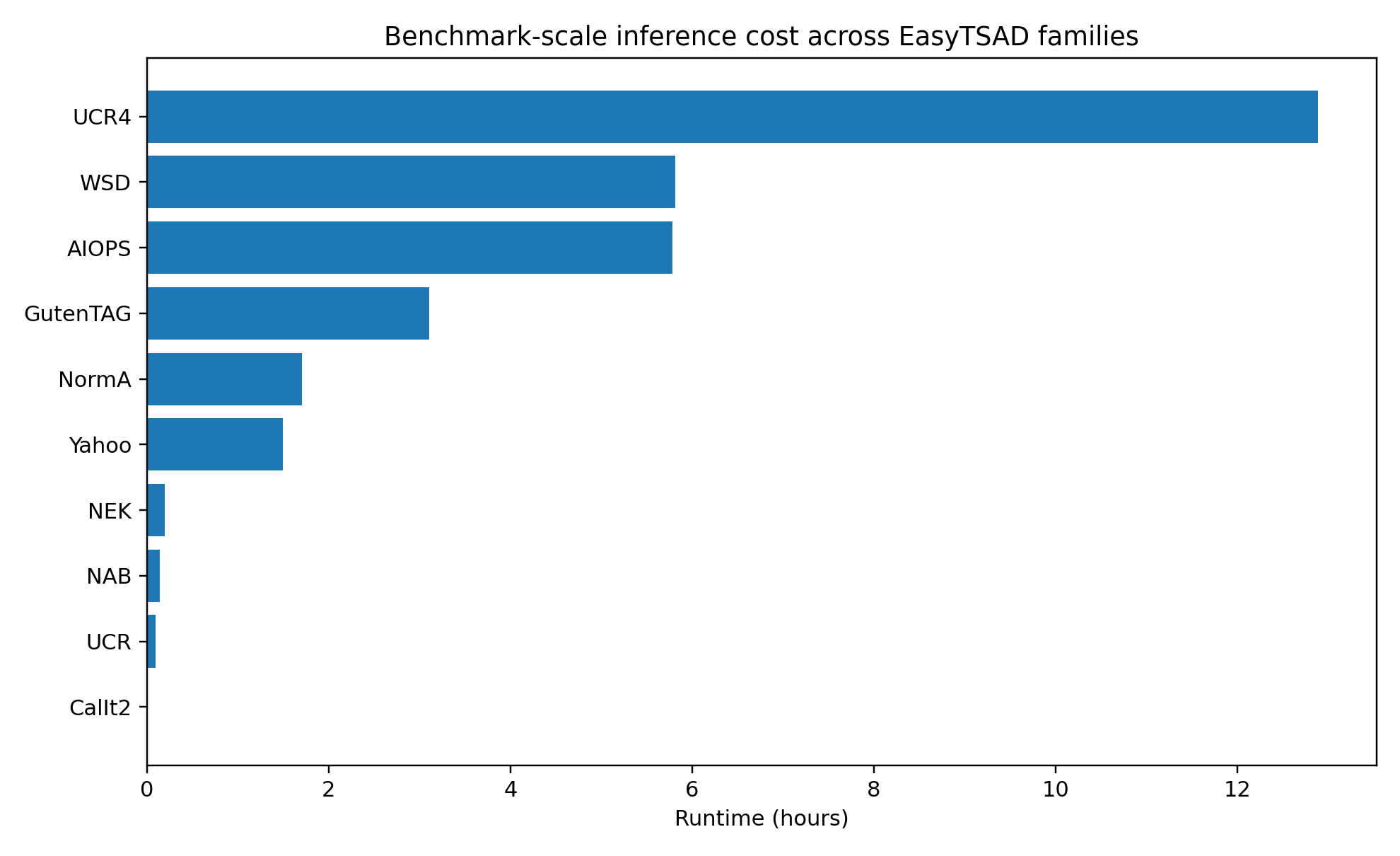}
    \caption{Benchmark-scale inference cost across EasyTSAD dataset families
    under the DDPM-based evaluation protocol. A small number of large families
    dominate the total wall-clock cost, particularly UCR4, AIOPS, WSD, and
    GutenTAG. This runtime concentration highlights the practical importance of
    accelerated samplers for large-scale validation and deployment of
    diffusion-based anomaly detection systems.}
    \label{fig:easytsad_runtime}
\end{figure}

The zero-shot benchmark results must be interpreted alongside their runtime cost.
Table~\ref{tab:easytsad_time} shows that full-suite EasyTSAD evaluation under
DDPM sampling requires 112{,}504.59 seconds ($\approx 31.25$ hours). Cost is
highly concentrated: UCR4 alone accounts for 12.89 hours, followed by WSD
(5.82 hours), AIOPS (5.78 hours), and GutenTAG (3.11 hours). Together, these
four families represent over 90\% of total wall-clock cost.
Figure~\ref{fig:easytsad_runtime} visualizes this concentration across all
families.

\begin{table}[H]
    \centering
    \caption{Zero-shot EasyTSAD inference runtimes under DDPM sampling
    (NV-Tesseract-AD, epoch~90 checkpoint). A small number of large families
    dominate wall-clock cost, motivating accelerated sampling for practical
    benchmark evaluation.}
    \label{tab:easytsad_time}
    \begin{tabularx}{\textwidth}{CCC}
        \toprule
        \textbf{Dataset Family} & \textbf{Test Time (s)} & \textbf{Test Time (h)} \\
        \midrule
        UCR      &      339.64 &  0.09 \\
        UCR4     & 46{,}388.58 & 12.89 \\
        AIOPS    & 20{,}824.84 &  5.78 \\
        NAB      &      525.11 &  0.15 \\
        Yahoo    &  5{,}387.37 &  1.50 \\
        WSD      & 20{,}935.09 &  5.82 \\
        NEK      &      717.80 &  0.20 \\
        GutenTAG & 11{,}197.46 &  3.11 \\
        NormA    &  6{,}155.75 &  1.71 \\
        CalIt2   &       32.95 &  0.01 \\
        \midrule
        \textbf{Total} & \textbf{112{,}504.59} & \textbf{31.25} \\
        \bottomrule
    \end{tabularx}
\end{table}

This runtime profile highlights a central practical constraint of diffusion-based
anomaly detection that is independent of model quality: even when the pretrained
backbone is fixed and no gradient computation is required, iterative sampling is
expensive enough to dominate benchmark-scale evaluation. The concentration of
cost in a small number of large families has a direct implication for
experimental design: studies that report results only on computationally cheap
families risk presenting an incomplete and potentially misleading picture of
performance. DPM-Solver++ addresses this constraint by reducing total suite
runtime by $3.5$--$5.7\times$ at the dataset level (approximately $40\times$
per window before parallelism effects), making comprehensive benchmark sweeps
computationally realistic. This changes not only the speed of individual
experiments but also the experimental scope that is feasible within a given
compute budget--enabling, for example, hyperparameter searches over solver
configurations or large-scale ablations that would be prohibitive under DDPM.

\subsection{Overall Interpretation}
\label{subsec:discussion}

Across all experiments, four conclusions emerge consistently and reinforce one
another.

\textbf{The dual-axis curriculum improves both reconstruction quality and masking
invariance.} The near-zero loss gap between complementary masking strategies after
the first curriculum stage confirms that staged easy-to-hard training regularizes
the denoiser toward a masking-invariant reconstruction regime without explicit
invariance constraints. This stability is a prerequisite for reliable residual-based
anomaly scoring, and its early emergence (within the first 30 epochs) validates
the theoretical motivation of the curriculum: well-conditioned early tasks
establish the representational foundation that generalizes across inference-time
corruption patterns.

\textbf{Train--inference alignment has an outsized effect on downstream detection.}
The $3.7\times$ F1 improvement between the non-curriculum baseline and
NV-Tesseract-AD is attributable jointly to curriculum-stabilized optimization and
consistent train--inference preprocessing. The two factors are mutually
reinforcing: curriculum training reduces residual variance (measurable as lower
training-phase loss and lower inference-time MAE/MSE), and consistent
preprocessing ensures that the resulting score distribution is preserved at test
time. Either factor in isolation produces a smaller gain. Normalization, clipping,
masking configuration, inference schedule, stage embedding, and checkpoint
selection must all be treated as integral components of the model specification,
not as separate implementation concerns.

\textbf{Thresholding is the dominant remaining detection bottleneck.} The gap
between default unsupervised SCS/MACS performance and the oracle AdjBestF1
analysis-only upper bound is substantially larger than the gap between the tuned
grid-search configuration and the oracle. This establishes that meaningful
headroom exists above default thresholding performance, and that the
reconstruction signal itself is not the limiting factor. Future gains will arise
primarily from stronger unsupervised calibration mechanisms--score normalization,
online threshold adaptation, and distribution-aware methods--rather than from
further scaling of the denoiser.

\textbf{Fast samplers are necessary for operational viability but must be
co-validated with the thresholding mechanism.} DPM-Solver++ reduces dataset-level
inference cost by $3.5$--$5.7\times$ and makes suite-level zero-shot evaluation
feasible within practical compute budgets. However, because solver choice alters
the residual score distribution, thresholding behavior changes with the sampler.
Any production deployment or comparative benchmark study must specify and validate
the sampler configuration as part of the full pipeline.

Taken together, these results support the central thesis of this work:
a curriculum-trained multivariate diffusion model can serve as a robust
zero-shot anomaly detector, but its practical effectiveness depends on three
tightly coupled components--stable reconstruction, calibrated thresholding,
and efficient sampling--that must be designed and validated as a single
integrated system. The strongest performance gains arise when all three
are aligned, and evaluations that treat any one component as a transparent
default risk substantially underestimating the model's true detection capability.

\section{Conclusions and Future Work}
\label{sec:conclusion}

We presented \textbf{NV-Tesseract-AD}, a curriculum-driven conditional diffusion
framework for multivariate time-series anomaly detection under missingness and
distributional shift. The core hypothesis motivating our approach--that
reconstruction difficulty can be explicitly structured along masking severity and
diffusion noise axes to stabilize optimization and improve residual
discriminability--is supported by three converging lines of evidence from our
experimental results.

First, the dual-axis curriculum over masking ratio and diffusion timestep produces
rapid and stable convergence: the diffusion noise-prediction loss gap between
complementary split-mask strategies collapses from 0.0263 at initialization to
effectively zero within the first curriculum stage and remains negligible
throughout training (Table~\ref{tab:recon_loss}), confirming that staged exposure
to increasing reconstruction difficulty leads to masking-invariant generalization
without explicit invariance constraints.

Second, curriculum training combined with consistent train--inference preprocessing
yields large gains in downstream anomaly detection relative to a non-curriculum
baseline ($\approx 3.7\times$ improvement in F1, $56\%$ reduction in MSE;
Table~\ref{tab:weights_comparison}). These results highlight that diffusion-based
anomaly detection quality depends jointly on model training and systems-level
pipeline alignment: small train--inference mismatches in normalization statistics,
masking parameters, inference schedule, or stage embedding configuration can
dominate end-to-end detection performance, motivating careful versioning of all
preprocessing and inference configuration as integral components of the model
specification.

Third, under a strict zero-shot protocol--training once on TSB-UAD-M and
evaluating across EasyTSAD families without any per-dataset tuning--NV-Tesseract-AD
transfers competitively on multiple benchmark families, with particularly strong
performance where anomaly structure aligns with reconstruction-residual scoring
(NAB F1 $= 0.9841$, NormA F1 $= 0.9477$; Table~\ref{tab:easytsad_f1}). These
results support reconstruction-centric diffusion training as a practical and
generalizable design principle for robust multivariate anomaly detection.

Despite these contributions, our results also surface two dominant remaining
bottlenecks that bound deployed performance. First, unsupervised threshold
calibration--necessary in the absence of anomaly labels--accounts for the
majority of the gap between model capability (as measured by the analysis-only
oracle AdjBestF1 bound) and deployed performance (as measured by SCS/MACS F1).
As shown in \S\ref{subsec:thresholding_results}, the grid-search analysis
demonstrates that this gap can be partially closed by hyperparameter tuning,
but the small margin between the best-tuned configuration (F1 $\approx 0.218$)
and the oracle (F1 $\approx 0.244$) confirms that the reconstruction signal
itself is not the limiting factor--it is the calibration mechanism. A systematic
approach to calibration-free thresholding under distribution shift therefore
remains an open and high-priority problem. Second, inference cost under ancestral
DDPM sampling ($\approx 31$ hours for the full EasyTSAD suite;
Table~\ref{tab:easytsad_time}) constrains practical deployability. DPM-Solver++
reduces this to dataset-level speedups of $3.5$--$5.7\times$ (approximately
$40\times$ per window before parallelism effects), but as shown in
\S\ref{subsec:sampler_results}, solver choice couples to thresholding behavior:
a sampler configuration that is optimal under DDPM may not remain optimal under
DPM-Solver++, and the two must be validated jointly.

Several directions follow naturally from this work. The most pressing is
unsupervised threshold calibration: our analysis identifies it as the dominant
remaining bottleneck, and methods that infer stable score geometry from the
reconstruction distribution without label access would directly translate to
deployed performance gains. Beyond calibration, detection-centric training
objectives--such as contrastive losses that explicitly widen the residual gap
between normal and anomalous inputs--could reduce reliance on post-hoc
thresholding entirely. Aligning the masking distribution between training and
inference, and jointly validating sampler choice with the thresholding mechanism,
are important systems-level steps for robust zero-shot deployment. Finally, a
theoretical account of why staged masking and noise schedules stabilize diffusion
optimization remains an open question worth pursuing.

\end{document}